\documentclass[10pt,twocolumn,letterpaper]{article}
\usepackage{amsmath}
\usepackage{lipsum}
\usepackage{indentfirst}
\usepackage{cases}
\usepackage{multirow} 
\usepackage{graphicx}
\usepackage{times}  
\usepackage{algorithm}
\usepackage{algorithmic}
\usepackage{color}
\usepackage{ulem}
\usepackage{pifont}
\usepackage{arydshln} 
\usepackage{float} 
\usepackage{bm}
\usepackage{caption}
\usepackage{graphicx}
\usepackage{amssymb}
\usepackage{booktabs}
\usepackage{cvpr}  

\usepackage[dvipsnames]{xcolor}

\definecolor{cvprblue}{rgb}{0.21,0.49,0.74}
\usepackage[pagebackref,breaklinks,colorlinks,citecolor=cvprblue]{hyperref}

\usepackage{pgfplots}
\usepackage{adjustbox}
\newcommand{\tht}[2]{\begin{tabular}{@{}#1@{}}#2\end{tabular}}

\usepackage[capitalize]{cleveref}
\crefname{section}{Sec.}{Secs.}
\Crefname{section}{Section}{Sections}
\Crefname{table}{Table}{Tables}
\crefname{table}{Tab.}{Tabs.}
\def\confName{CVPR}
\def\confYear{2024}

\title{
SSR-Encoder: Encoding Selective Subject Representation for Subject-Driven Generation
}

\author{
Yuxuan Zhang\textsuperscript{\rm 1*\footnotemark[0]},
Yiren Song\textsuperscript{\rm 4},
Jiaming Liu\textsuperscript{\rm 2}$^{\dagger}$,
Rui Wang\textsuperscript{\rm 3*},
Jinpeng Yu\textsuperscript{\rm 6*}, 
Hao Tang\textsuperscript{\rm 5}, \\
Huaxia Li\textsuperscript{\rm 2},
Xu Tang\textsuperscript{\rm 2},
Yao Hu\textsuperscript{\rm 2},
Han Pan\textsuperscript{\rm 1},
Zhongliang Jing\textsuperscript{\rm 1}$^{\dagger}$
}

\begin{document}

\twocolumn[{%
\maketitle
\centering
\vspace{-0.7cm}
\noindent
\normalsize\textsuperscript{\rm 1}Shanghai Jiao Tong University, \textsuperscript{\rm 2}Xiaohongshu Inc., \textsuperscript{\rm 3}Beijing University of Posts and Telecommunications, \\
\textsuperscript{\rm 4}National University of Singapore,
\textsuperscript{\rm 5}Carnegie Mellon University, \textsuperscript{\rm 6}ShanghaiTech University\\


\begin{figure}[H]
\hsize=\textwidth 
\centering
\includegraphics[width=0.99\textwidth]{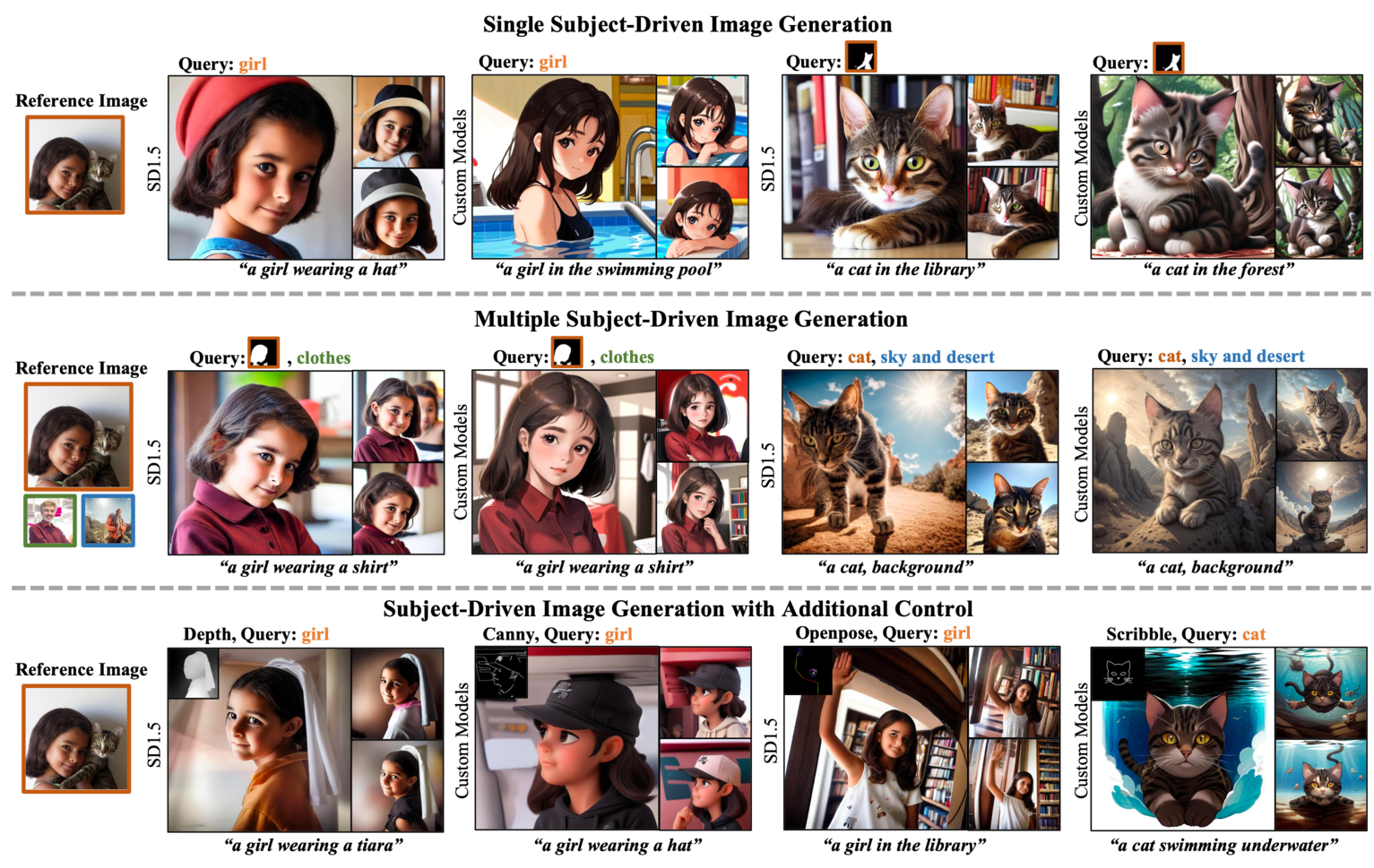}
\caption{Our \textit{SSR-Encoder} is a model generalizable encoder, which is able to guide any customized diffusion models for single subject-driven image generation (top branch) or multiple subject-driven image generation from different images (middle branch) based on the image representation selected by the text query or mask query without any additional test-time finetuning. Furthermore, our \textit{SSR-Encoder} can also be applied for the controllable generation with additional control (bottom branch).}
    \label{teaser}
\end{figure}
}]

\renewcommand{\thefootnote}{\fnsymbol{footnote}}
\footnotetext[1]{Work done during internship at Xiaohongshu Inc.}
\footnotetext[2]{Corresponding authors.}
\renewcommand{\thefootnote}{\arabic{footnote}}

\begin{abstract}
Recent advancements in subject-driven image generation have led to zero-shot generation, yet precise selection and focus on crucial subject representations remain challenging. Addressing this, we introduce the SSR-Encoder, a novel architecture designed for selectively capturing any subject from single or multiple reference images. It responds to various query modalities including text and masks, without necessitating test-time fine-tuning. The SSR-Encoder combines a Token-to-Patch Aligner that aligns query inputs with image patches and a Detail-Preserving Subject Encoder for extracting and preserving fine features of the subjects, thereby generating subject embeddings. These embeddings, used in conjunction with original text embeddings, condition the generation process. Characterized by its model generalizability and efficiency, the SSR-Encoder adapts to a range of custom models and control modules. Enhanced by the Embedding Consistency Regularization Loss for improved training, our extensive experiments demonstrate its effectiveness in versatile and high-quality image generation, indicating its broad applicability. Project page: \url{ssr-encoder.github.io}
\end{abstract} 
\section{Introduction}
\label{sec:intro}

Recent advancements in image generation, especially with the advent of text-to-image diffusion models trained on extensive datasets, have revolutionized this field. A prime example is Stable Diffusion, an open-source model cited as \cite{ldm}, which allows a broad user base to easily generate images from textual prompts. A growing area of interest that has emerged is the subject-driven generation, where the focus shifts from creating a generic subject, like ``a cat'' to generating a specific instance, such as ``the cat''. However, crafting the perfect text prompt to generate the desired subject content poses a significant challenge. Consequently, researchers are exploring various strategies for effective subject-driven generation.

Subject-driven image generation aims to learn subjects from reference images and generate images aligning with specific concepts like identity and style. Currently, one prominent approach involves test-time fine-tuning \cite{TI, DB,customdiffusion,breakascene}, which, while efficient, requires substantial computational resources to learn each new subject.
Another approach \cite{e4t, taming, instantbooth, chen2023anydoor,elite} encodes the reference image into an image embedding to bypass the fine-tuning cost. However, these encoder-based models typically require joint training with the base diffusion model, limiting their generalizability. A concurrent work, IP-adapter \cite{ye2023ip}, tackles both fine-tuning costs and generalizability by learning a projection to inject image information into the U-Net, avoiding the need to fine-tune the base text-to-image model, thereby broadening its application in personalized models.

Despite these advancements, a critical aspect often overlooked is the extraction of the most informative representation of a subject. With images being a complex mixture of subjects, backgrounds, and styles, it's vital to focus on the most crucial elements to represent a subject effectively. To address this, we introduce the \textit{SSR-Encoder}, an image encoder that generates \textbf{S}elective \textbf{S}ubject \textbf{R}epresentations for subject-driven image generation. 

Our \textit{SSR-Encoder} firstly aligns patch-level visual embeddings with texts in a learnable manner, capturing detailed subject embeddings guided by token-to-patch attention maps. Furthermore, we propose subject-conditioned generation, which utilizes trainable copies of cross-attention layers to inject multi-scale subject information. A novel Embedding Consistency Regularization Loss is proposed to enhance the alignment between text queries and visual representations in our subject embedding space during training. This approach not only ensures effective token-to-patch alignment but also allows for flexible subject selection through text and mask queries during inference. Our \textit{SSR-Encoder} can be seamlessly integrated into any customized stable diffusion models without extensive fine-tuning. Moreover, the \textit{SSR-Encoder} is adaptable for controllable generation with various additional controls, as illustrated in Fig. \ref{teaser}.

We summarize our main contributions as follows:
\begin{itemize}
    \item We propose a novel framework, termed as \textit{SSR-Encoder}, for selective subject-driven image generation. It allows selective single- or multiple-subject generation, fully compatible with ControlNets (e.g.\ canny, OpenPose, etc.), and customized stable diffusion models without extra test-time training. 
    \item Token-to-Patch Aligner and Detail-Preserved Subject Encoder are proposed in our \textit{SSR-Encoder} to learn selective subject embedding. We also present an Embedding Consistency Regularization Loss to enhance token-to-patch text-image alignment in the subject embedding space.  
    \item Our extensive experiments have validated the robustness and flexibility of our approach, showcasing its capability to deliver state-of-the-art (SOTA) results among finetuning-free methods. Impressively, it also demonstrates competitive performance when compared with finetuning-based methods. 
\end{itemize}
\section{Related Work}
\label{sec:related}

\textbf{Text-to-image diffusion models.}
In recent years, text-to-image diffusion models \cite{dalle, dalle2,ldm, Imagen, IF, sdxl, raphael, ediffi, glide, zeng2023ipdreamer} have made remarkable progress, particularly with the advent of diffusion models, which have propelled text-to-image generation to large-scale commercialization. DALLE \cite{dalle} first achieved stunning image generation results using an autoregressive model. Subsequently, DALLE2 \cite{dalle2} employed a diffusion model as the generative model, further enhancing text-to-image synthesis ability. Imagen \cite{Imagen} and Stable Diffusion \cite{ldm} trained diffusion models on larger datasets, further advancing the development of diffusion models and becoming the mainstream for image generation large models. DeepFloyd IF \cite{IF} utilized a triple-cascade diffusion model, significantly enhancing the text-to-image generation capability, and even generating correct fonts. Stable Diffusion XL \cite{sdxl}, a two-stage cascade diffusion model, is the latest optimized version of stable diffusion, greatly improving the generation of high-frequency details, small object features, and overall image color.

\textbf{Controllable image generation.}
Current diffusion models can incorporate additional modules, enabling image generation guided by multimodal image information such as edges, depth maps, and segmentation maps. These multimodal inputs significantly enhance the controllability of the diffusion model's image generation process. Methods like ControlNet \cite{controlnet} utilize a duplicate U-Net structure with trainable parameters while keeping the original U-Net parameters static, facilitating controllable generation with other modal information. T2I-adapter \cite{t2i} employs a lightweight adapter for controlling layout and style using different modal images. Uni-ControlNet \cite{unicontrol} differentiates between local and global control conditions, employing separate modules for injecting these control inputs. Paint by Example \cite{paintbyexample} allows for specific region editing based on reference images. Other methods \cite{attendandexcite, realworld, masactrl, nullTI, p2p,selfguidance} manipulate the attention layer in the diffusion model's denoising U-Net to direct the generation process. P2P \cite{p2p} and Null Text Inversion \cite{nullTI} adjust cross-attention maps to preserve image layout under varying text prompts.

\textbf{Subject-driven image generation.}
Subject-driven image generation methods generally fall into two categories: those requiring test-time finetuning and those that do not. The differences in characteristics of these methods are illustrated in Table \ref{tab:my_label}. Test-time finetuning methods \cite{DB, TI, customdiffusion, svdiff, lora, breakascene,disenbooth,p+,designaencoder} often optimize additional text embeddings or directly fine-tune the model to fit the desired subject. For instance, Textual Inversion \cite{TI} optimizes additional text embeddings, whereas DreamBooth \cite{DB} adjusts the entire U-Net in the diffusion model. Other methods like Customdiffusion \cite{customdiffusion} and SVDiff \cite{svdiff} minimize the parameters needing finetuning, reducing computational demands. Finetuning-free methods \cite{instantbooth,e4t,face0, elite,ye2023ip, taming, blipdiffusion,subject} typically train an additional structure to encode the reference image into embeddings or image prompts without additional finetuning. ELITE \cite{elite} proposes global and local mapping training schemes to generate subject-driven images but lack fidelity. Instantbooth \cite{instantbooth} proposes an adapter structure inserted in the U-Net and trained on domain-specific data to achieve domain-specific subject-driven image generation without finetuning. IP-adapter \cite{ye2023ip} encodes images into prompts for subject-driven generation. BLIP-Diffusion \cite{blipdiffusion} enables efficient finetuning or zero-shot setups. However, many of these methods either utilize all information from a single image, leading to ambiguous subject representation, or require finetuning, limiting generalizability and increasing time consumption. In contrast, our SSR-Encoder is both generalizable and efficient, guiding any customized diffusion model to generate images based on the representations selected by query inputs without any test-time finetuning.

\begin{table}[!tbp]
    \centering
    \footnotesize{
      \caption{Comparative Analysis of Previous works. Considering Fine-Tuning free, Model Generalizability, and Selective Representation, SSR-Encoder is the first method offering all three features.}
    \label{tab:my_label}
    \setlength{\tabcolsep}{1mm} 
    \begin{tabular}{c|c|c|c} \toprule 
Method & \tht{c}{Finetuning\\-free} & \tht{c}{Model \\ Generalizable} & \tht{c}{Selective\\Representation} \\  \midrule 

Textual Inversion \cite{TI} & \ding{55} & \ding{51} & \ding{55} \\ \hline
Dreambooth \cite{DB} & \ding{55} & \ding{55} & \ding{55} \\ \hline
LoRA \cite{lora} & \ding{55} & \ding{51} & \ding{55} \\ \hline
Custom diffusion \cite{customdiffusion} & \ding{55} & \ding{55} & \ding{55} \\ \hline
Break-A-Scene \cite{breakascene} & \ding{55} & \ding{55}  & \ding{51} \\ \hline
E4T \cite{e4t} & \ding{55} & \ding{55}  & \ding{55} \\ \hline
Instantbooth \cite{instantbooth} & \ding{51} & \ding{55} & \ding{55} \\ \hline
ELITE \cite{elite} & \ding{51} & \ding{55} & \ding{55} \\ \hline
Taming \cite{taming} & \ding{51} & \ding{55} & \ding{55} \\ \hline
IP-adapter \cite{ye2023ip} & \ding{51} & \ding{51} & \ding{55} \\ \hline
BLIP-diffusion \cite{blipdiffusion} & \ding{51} & \ding{55}& \ding{51} \\ \hline
SSR-Encoder(Ours) & \ding{51} & \ding{51}  & \ding{51} \\ \bottomrule
    \end{tabular}
    }
    
\end{table}
\section{The Proposed Method}
\label{sec:method}
\begin{figure*}[!h]
    \centering
    \vspace{-0.8cm}
\includegraphics[width=1.0\linewidth]{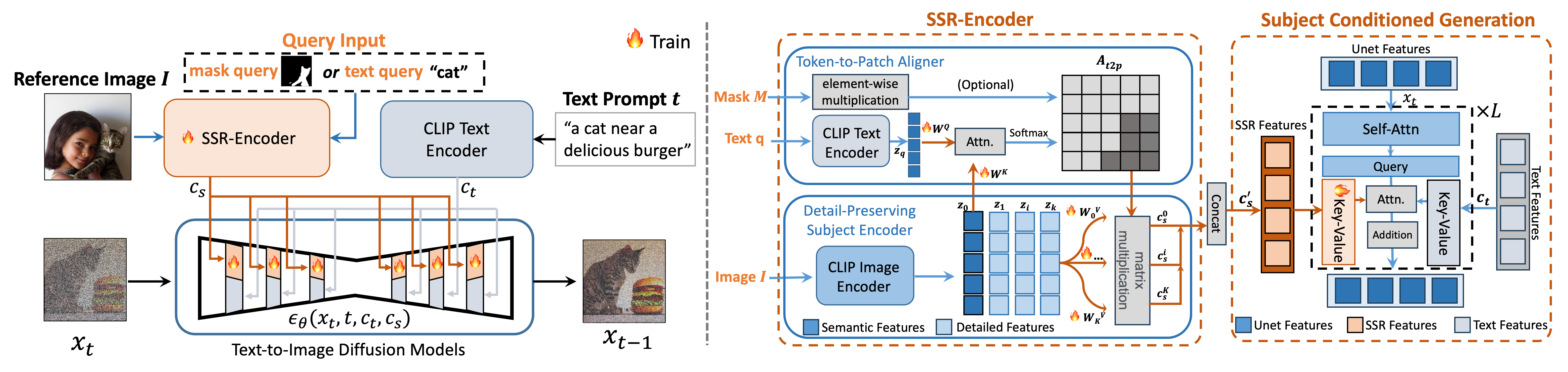}
    \caption{Overall schematics of our method. 
    Given a query text-image pairs $\left(q, I\right)$, the SSR-Encoder employs a token-to-patch aligner to highlight the selective regions in the reference image by the query. 
    It extracts more fine-grained details of the subject through the detail-preserving subject encoder, projecting multi-scale visual embeddings via the token-to-patch aligner.
    Then, we adopt subject-conditioned generation to generate specific subjects with high fidelity and creative editability.
    During training, we adopt reconstruction loss $L_{LDM}$ and embedding consistency regularization loss $L_{reg}$ for selective subject-driven learning.}
    \label{method}
    \vspace{-0.5cm}
\end{figure*}

Selective subject-driven image generation aims to generate target subjects in a reference image with high fidelity and creative editability, guided by the user's specific queries (text or mask). To tackle this, we propose our SSR-Encoder, a specialized framework designed to integrate with any custom diffusion model without necessitating test-time fine-tuning.

Formally, for a given reference image $\mathit{I}$ and a user query $\mathit{q}$, the SSR-Encoder effectively captures subject-specific information and generates multi-scale subject embeddings $\mathit{c_s}$. These multi-scale subject embeddings $\mathit{c_s}$ are subsequently integrated into the U-Net model with trainable copies of cross-attention layers. The generation process, conditioned on both subject embeddings $c_s$ and text embedding $c_t$, allows for the production of desired subjects with high fidelity and creative editability. The overall methodology is illustrated in Fig. \ref{method}.

In general, SSR-Encoder is built on text-to-image diffusion models~\cite{ldm}\footnote{Reviewed in the Supplementary.}. It comprises two key components: the token-to-patch aligner and detail-preserving subject encoder (Sec. \ref{sec:3.2}). The subject-conditioned generation process is detailed in Sec. \ref{sec:3.3}. Lastly, training strategies and loss functions are presented in Sec. \ref{sec:3.4}.

\subsection{Selective Subject Representation Encoder}
\label{sec:3.2}

Our Selective Subject Representation Encoder (SSR-Encoder) is composed of two integral parts: Token-to-Patch Aligner and Detail-Preserving Subject Encoder. The details of each component are as follows.

\textbf{Token-to-patch aligner}. Several works \cite{maskclip, clipsurgery, dinoregister} have pointed out that CLIP tends to prioritize background regions over foreground subjects when identifying target categories. Therefore, relying solely on text-image similarity may not adequately capture subject-specific information. To address this issue, we propose the Token-to-Patch (T2P) Aligner, which implements two trainable linear projections to align image patch features with given text token features. Mathematically, given a query text-image pair $(\mathit{q}, \mathit{I})$, we employ pre-trained CLIP encoders to generate the text query and image reference into query embedding $\mathit{z_q}\in \mathbb{R}^{N_q \times D_q}$ and semantic visual embedding $\mathit{z_0}\in \mathbb{R}^{N_i \times D_i}$ from the last CLIP layer, respectively, where $N_{\left(\cdot\right)}$ and $D_{\left(\cdot\right)}$ represent the number of tokens and dimensions for query and image features respectively. We then use the trainable projection layers $\mathbf{W^Q}$ and $\mathbf{W^K}$ to transform them into a well-aligned space. The alignment is illustrated as follows:

\begin{equation}
\begin{aligned}
\label{equal:aligner}
\mathit{Q}&=\mathbf{W^Q}\cdot \mathit{z_q}, \\ \mathit{K}&=\mathbf{W^K}\cdot \mathit{z_0},
\end{aligned}
\end{equation}

\begin{equation}
\label{equal:attn_operation}
\mathit{A_{t2p}}=\operatorname{Softmax}\left(\frac{\mathit{Q} \mathit{K^{\top}}}{\sqrt{d}}\right),
\end{equation}
where $\mathit{A_{t2p}} \in \mathbb{R}^{N_t \times N_i}$ represents the token-to-patch attention map.

Furthermore, the $\mathit{A_{t2p}}$ matrix serves a dual purpose: similarity identification and region selection. Consequently, our aligner naturally supports \textbf{mask-based query}. In practice, we can manually assign a mask $\mathit{M}$ to $\mathit{A_{t2p}}$ for mask-guided generation with null-text query inputs. Following Eq.~\eqref{equal:attn_operation}, we can proceed to reweight $\mathit{A_{t2p}}$ using the predefined mask $\mathit{M}$ to highlight selected regions, ensuring our SSR-Encoder focuses solely on the selected valid regions of reference images.

\textbf{Detail-preserving subject encoder}. 
Following most of the preceding methods~\cite{ye2023ip, pada, elite}, we employ a pre-trained CLIP visual backbone to extract image representations from reference images. However, the conventional practice of extracting visual embeddings $z_0$ from the last CLIP layer does not align with our objective of preserving fine details to the maximum extent. Our preliminary experiments\footnote{Detailed in the supplementary.} have identified a notable loss of fine-grained details in the semantic image features $z_0$. Addressing this, we introduce the detail-preserving subject encoder, which extracts features across various layers to preserve more fine-grained details. Formally, the visual backbone processes an image $\mathit{I}$ to produce multi-scale detailed image features $\mathit{z_I} = \{\mathit{z_k}\}_{k=0}^{K}$, where $\mathit{z_0}$ represents semantic visual embedding used in T2P aligner and $\mathit{z_k}$ represents other detailed visual embeddings at the scale of $k$ in CLIP visual backbone and $K$ refers to the number of target scales. We set $K$ to 6 in all experimental settings.

To fully leverage the benefits of multi-scale representation, we adopt separate linear projections $\mathbf{W_k^V}$ for image feature $\mathit{z_k}$ at different scales. Combining with the token-to-patch attention map $\mathit{A_{t2p}}$, the subject embeddings $\mathit{c_s}=\{\mathit{c_s^k}\}^K_{k=0}$ are computed as per Eq. \eqref{subject_projection}:

\begin{equation}
\label{subject_projection}
\mathit{V_k}=\mathbf{W^V_k}\cdot \mathit{z_k}, \\
\mathit{c_s^k}={\mathit{A_{t2p}} \mathit{V^{\top}_k}},
\end{equation}
where $\mathit{c^k_s}$ denotes subject embedding at scale of~$k$.
Our SSR-Encoder now enables to capture multi-scale subject representation $\mathit{c_s}=\{\mathit{c_s^k}\}_{k=0}^K$, which are subsequently used for subject-driven image generation via subject-conditioned generation process.

\subsection{Subject Conditioned Generation}
\label{sec:3.3}
In our approach, $\mathit{c_s}$ is strategically projected into the cross-attention layers of the U-Net. This is achieved through newly added parallel subject cross-attention layers, each corresponding to a text cross-attention layer in the original U-Net. Rather than disturbing the text embedding $\mathit{c_t}$, these new layers independently aggregate subject embeddings $\mathit{c_s}$. Inspired by works like \cite{ye2023ip,customdiffusion, controlnet, elite}, we employ trainable copies of the text cross-attention layers to preserve the efficacy of the original model. The key and value projection layers are then adapted to train specifically for a subject-conditioned generation. 
To full exploit of both global and local subject representation, we concatenate all $\mathit{c_s^k}$ at the token dimension before projection, i.e. $\mathit{c_s^{\prime}}=\operatorname{concat}\left( \mathit{c_s^k}, \operatorname{dim}=0\right)$, where $\mathit{c_s^k}\in \mathbb{R}^{N_q\times D_i}$ represents subject representation at the scale of $k$. 
The output value $\mathit{O}$ of the attention layer is formulated as follows:

\vspace{-0.3cm}
\begin{equation}
\begin{split}
\mathit{O}
&=\underbrace{\operatorname{CrossAttention}\left(\mathbf{Q}, \mathbf{K}, \mathbf{V}, \mathit{c_t}, \mathit{x_t}\right)}_{\textrm{text condition}} \\
&+\lambda\underbrace{\operatorname{CrossAttention}\left(\mathbf{Q}, \mathbf{K_{S}}, \mathbf{V_{S}}, \mathit{c_s^{\prime}}, \mathit{x_t}\right)}_{\textrm{subject condition}},
\end{split}
\end{equation}
where $\mathit{c_t}$ represents the text embedding and $\mathit{x_t}$ represents the latent.
$\mathbf{Q}, \mathbf{K}, \mathbf{V}$ represents query, key, and value projection layers in the original text branch respectively while $\mathbf{K_{S}}, \mathbf{V_{S}}$ represents trainable copies of key and value projection layers for concatenated subject embedding $\mathit{c_s}$. $\lambda$ is a weight adjustment factor, with a default value of 1.

By our subject-conditioned generation, text-to-image diffusion models can generate target subjects conditioned on both text embeddings and subject embeddings. 

\subsection{Model Training and Inference}
\label{sec:3.4}
During the training phase, our model processes paired images and texts from multimodal datasets. The trainable components include the token-to-patch aligner and the subject cross-attention layers.

In contrast to CLIP, which aligns global image features with global text features, our token-to-patch aligner demands a more granular token-to-patch alignment. To achieve this, we introduce an Embedding Consistency Regularization Loss $L_{reg}$. This loss is designed to enhance similarity between the subject embeddings $\mathit{c_s}$ and the corresponding query text embedding $\mathit{z_q}$, employing a cosine similarity function as demonstrated in Eq. \eqref{equ:reg}:

\begin{equation}
\begin{aligned}
\mathit{\overline{c_{s}}} &= \operatorname{Mean}\left(
\mathit{c_{s}^0}, \mathit{c_{s}^1}, ..., \mathit{c_{s}^K} \right), \\
\mathcal{L}_{reg} &= \operatorname{Cos} \left(\mathit{\overline{c_{s}}}, \mathit{z_q}\right) = 1 - \frac{\mathit{\overline{c_{s}}} \cdot \mathit{z_q}}{|\mathit{\overline{c_{s}}}||\mathit{z_q}|},
\label{equ:reg}
\end{aligned}
\end{equation}
where $\overline{\mathit{c_s}}$ is the mean of subject embeddings and $\mathit{z_q}$ represents the query text embeddings. 
As illustrated in Fig.~\ref{image8}, our T2P Aligner, trained on a large scale of image-text pairs, can effectively align query text with corresponding image regions. This capability is a key aspect of selective subject-driven generation.

Similar to the original Stable diffusion model, our training objective also includes the same $\mathcal{L}_{LDM}$ loss, as outlined in Eq. \eqref{equ:simple}:
\begin{equation}
\mathcal{L}_{LDM}(\bm{\theta})=\mathbb{E}_{\mathit{x_0}, t, \epsilon}\left[\left\|\epsilon-\epsilon _{\bm{\theta}}\left(\mathit{x_t}, t, \mathit{c_t}, \mathit{c_s}\right)\right\|_2^2\right],
\label{equ:simple}
\end{equation}
where $\mathbf{x_t}$ is the noisy latent at time step $t$, $\epsilon$ is the ground-truth latent noise. $\epsilon_{\bm{\theta}}$ is the noise prediction model with parameters $\bm{\theta}$.

Thus, our total loss function is formulated as:
\begin{equation}
\mathcal{L}_{total}=\mathcal{L}_{LDM} + \tau \mathcal{L}_{reg},
\end{equation}
where $\tau$ is set as a constant, with a value of 0.01. As depicted in Fig. \ref{image4} (in the last column), the inclusion of $\mathcal{L}_{reg}$ significantly enhances the text-image alignment capabilities of the SSR-Encoder. This improvement is evident in the generated images, which consistently align with both the subject prompt and the details of the reference image.

During inference, our method has the ability to decompose different subjects from a single image or multiple images. By extracting separate subject embeddings for each image and concatenating them together, our SSR-Encoder can seamlessly blend elements from multiple scenes. This flexibility allows for the creation of composite images with high fidelity and creative versatility.
\section{Experiment}
\label{sec:exper}

\begin{figure*}[!hp]
\centering
\vspace{-0.7cm}
\includegraphics[width=0.92\textwidth]{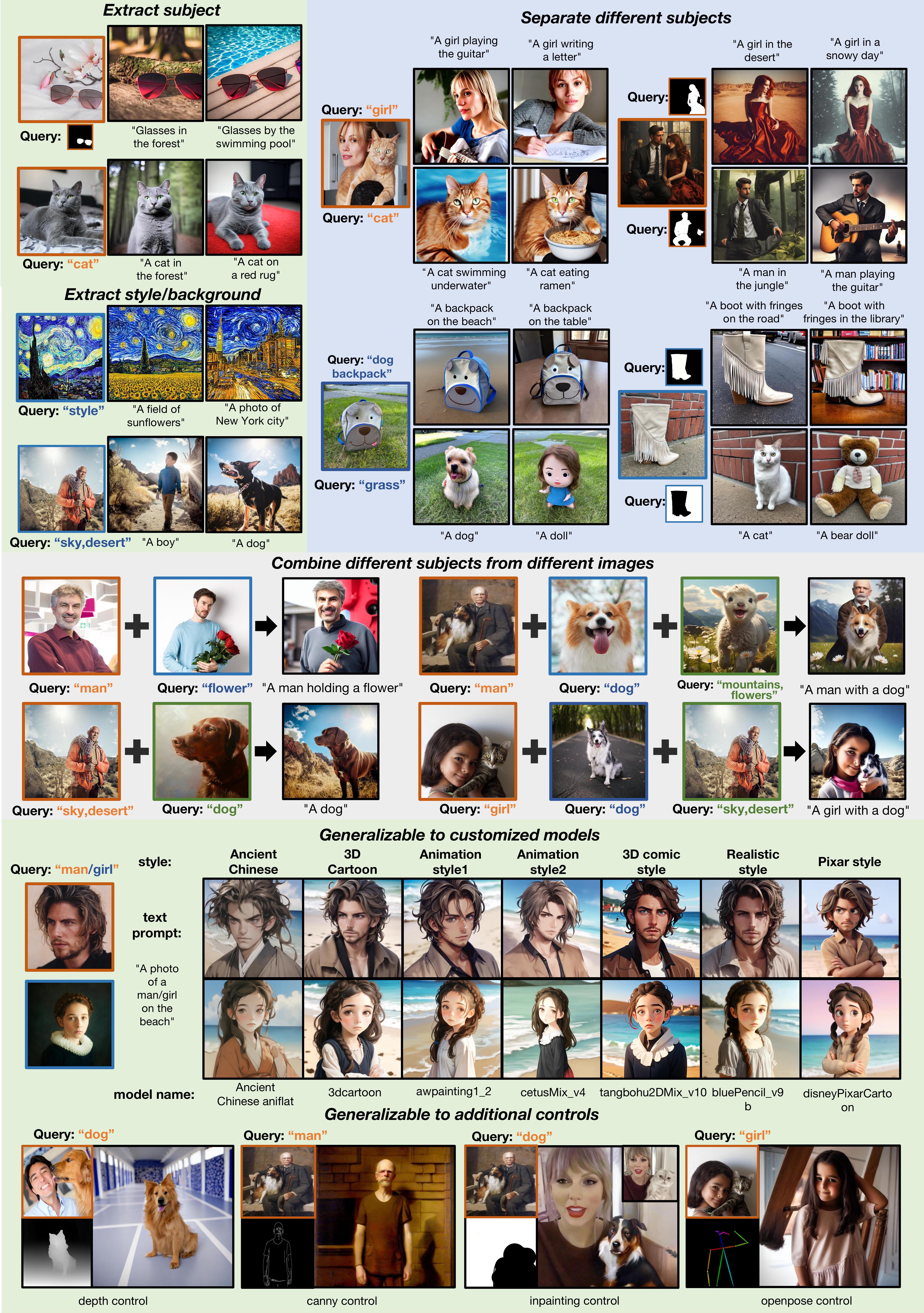}
    \caption{\textbf{Qualitative results of SSR-Encoder} in different generative capabilities. Our method supports two query modalities and is adaptable for a variety of tasks, including single- and multi-subject conditioned generation. Its versatility extends to integration with other customized models and compatibility with off-the-shelf ControlNets.}
    \label{image1}
\end{figure*}

\begin{figure*}[!h]
     \centering   
     \vspace{-1cm}
\includegraphics[width=1\textwidth]{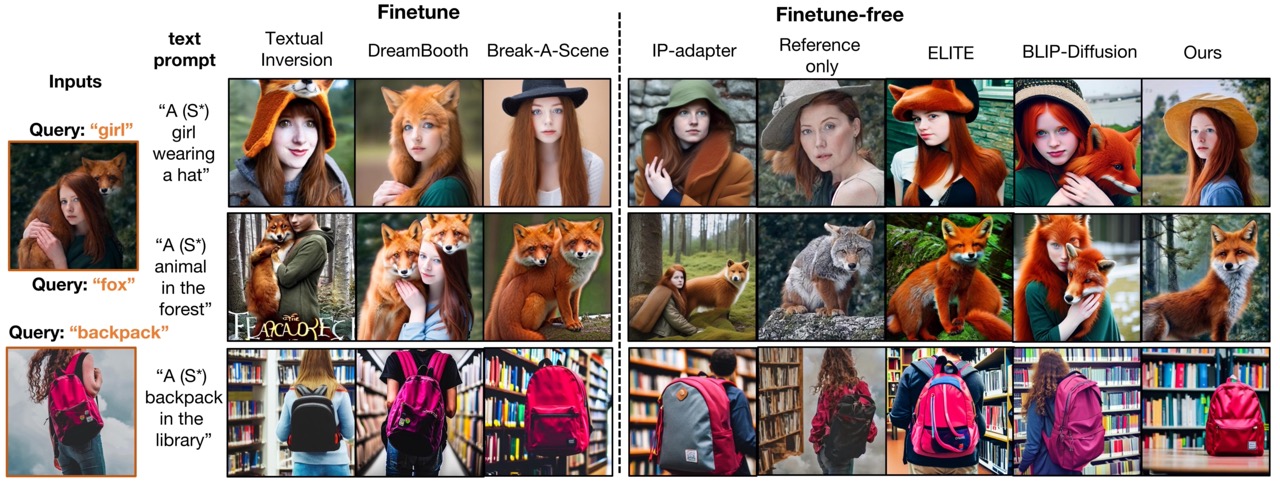}
\caption{\textbf{Qualitative comparison} of different methods. Our results not only excel in editability and exclusivity but also closely resemble the reference subjects in visual fidelity. Notably, the SSR-Encoder achieves this without the need for fine-tuning. }
\vspace{-0.5cm}
\label{compare}
\end{figure*}

\subsection{Experimental Setup}
\textbf{Training data.} Our model utilizes the Laion 5B dataset, selecting images with aesthetic scores above 6.0. The text prompts are re-captioned using BLIP2. The dataset comprises 10 million high-quality image-text pairs, with 5,000 images reserved for testing and the remainder for training.

\textbf{Implementation details.} We employed Stable Diffusion V1-5 as the pre-trained diffusion model, complemented by the pre-trained CLIP text encoder. For training, images are resized to ensure the shortest side is 512 pixels, followed by a center crop to achieve a 512$\times$512 resolution, and sent to the stable diffusion. The same image is resized to 224$\times$224 and sent to the SSR encoder. The model training process is divided into two steps. In the first step, the multi-scale strategy is not employed, and the model is trained for 1 million steps on 8H800s GPUs, with a batch size of 16 per GPU and a learning rate of 5e-5. In the second step, the same hyper-parameters are used, and the model parameters obtained from the first step are used as the initialization parameters. The multi-scale strategy is employed in this step to train the model for an additional 100,000 steps. Inference was performed using DDIM as the sampler, with a step size of 30 and a guidance scale set to 7.5. 
\subsection{Evaluation Metrics}
To evaluate our model, we employ several metrics and datasets:
\begin{itemize}
    \item \textbf{Multi-subject bench}: We created a benchmark with 100 images, each containing 2-3 subjects.
    \item \textbf{DreamBench datasets} \cite{DB}: This dataset includes 30 subjects, each represented by 4-7 images.
\end{itemize}

For a comprehensive comparison with state-of-the-art (SOTA) methods, we employed the following metrics: \textbf{DINO Scores}\cite{dinov1}, \textbf{CLIP-I}\cite{clip} and \textbf{DINO-M Scores} to assess subject alignment, \textbf{CLIP-T} \cite{clipscore} for evaluating image-text alignment, \textbf{CLIP Exclusive Score (CLIP-ES)} to measure the exclusivity of subject representation, and the \textbf{Aesthetic Score} \cite{laion5b} to gauge the overall quality of the generated images. 

Notably, CLIP-ES is calculated by generating an image $I$ using prompts for subject $A$ from a reference image and evaluating the CLIP-T score with a different subject $B$ and $I$. A lower CLIP-ES score indicates higher exclusivity. The DINO-M score, specifically designed for multiple subjects, evaluates identity similarity between masked versions of input and generated images, as detailed in \cite{breakascene}. Both CLIP-ES and DINO-M scores are evaluated on the Multi-Subject Bench.

\subsection{Comparison Methods}
For a comprehensive evaluation of our method, we benchmarked it against a range of state-of-the-art (SOTA) techniques. The methods we compared are categorized based on their approach to fine-tuning. In the fine-tuning-based category, we include \textbf{Textual Inversion} \cite{TI}, \textbf{Dreambooth} \cite{DB}, and \textbf{Break-a-Scene} \cite{breakascene}. For fine-tuning-free methods, our comparison encompassed \textbf{Reference Only} \cite{Mikubill2022}, \textbf{Elite} \cite{elite}, \textbf{IP-adapter} \cite{ye2023ip}, and \textbf{BLIPDiffusion} \cite{blipdiffusion}. This selection of methods provides a diverse range of approaches for a thorough comparative analysis with our SSR-Encoder.

\subsection{Experiment Results}

\setlength{\tabcolsep}{2.6mm}{
\begin{table*}[!h]
\begin{footnotesize}
\centering
\caption{\textbf{Quantitative comparison} of different methods. Metrics that are bold and underlined represent methods that rank 1st and 2nd, respectively. $^{\dagger}$ indicates that the experimental value is referenced from BLIP-Diffusion\cite{blipdiffusion}.}
\label{tab2}
\begin{tabular}{*{2} cc|ccc|ccc|c}
\toprule
& \multirow{2}*{\small Type} & \multirow{2}*{\small Method} & \small CLIP-T $\uparrow$ & \small CLIP-ES $\downarrow$  & \small DINO-M $\uparrow$ & \small CLIP-T $\uparrow$ & \small DINO $\uparrow$ & \small CLIP-I $\uparrow$ & \small Aesthetic  \\
& & &\multicolumn{3}{c}{\small (Multi-subject bench)} &\multicolumn{3}{c}{\small (DreamBench)} & Score$\uparrow$ \\
\midrule
 &   &Textual Inversion &0.240 & 0.212 & 0.410 &0.255$^{\dagger}$ &0.569$^{\dagger}$ &0.780$^{\dagger}$ & 6.029\\
 & Finetune-based  &Dreambooth & \underline{0.298} & 0.223 &\textbf{0.681} &\underline{0.305}$^{\dagger}$ &\textbf{0.668}$^{\dagger}$ & \underline{0.803}$^{\dagger}$ & \underline{6.330} \\
 & methods &Break-A-Scene &0.285 & \underline{0.187} & \underline{0.630} &0.287 &\underline{0.653} &\underline{0.788} & 6.234 \\
 \cdashline{3-10}
 & &Ours(full) &\textbf{0.302} & \textbf{0.182} &0.556 &\textbf{0.308} &0.612 &\textbf{0.821} &\textbf{6.563} \\
 \hline{}
 
 &   &BLIP-Diffusion &\underline{0.287} & 0.198 & \underline{0.514} &\underline{0.300}$^{\dagger}$ &0.594$^{\dagger}$ &0.779$^{\dagger}$ & 6.212 \\
 &   &Reference only &0.242 & 0.195 & 0.434 &0.286  &0.542 &0.727 & 5.812 \\
 & Finetune-free  &IP-adapter &0.272 & 0.201 & 0.442 &0.274 &\underline{0.608} &\underline{0.809} & \underline{6.432} \\
 & methods &ELITE &0.253 & \underline{0.194} & 0.483 &0.298 &0.605 &0.775 & 6.283 \\
 \cdashline{3-10}
 &  &Ours(full) &\textbf{0.302}& \textbf{0.182} &\textbf{0.556} &\textbf{0.308} & \textbf{0.612} &\textbf{0.821} &\textbf{6.563} \\
\bottomrule
\end{tabular}
\end{footnotesize}
\vspace{-0.7cm}
\end{table*}
}

\textbf{Quantitative comparison}. 
Table~\ref{tab2} presents our quantitative evaluation across two benchmarks: the Multi-Subject Bench and DreamBench. Overall, SSR-Encoder clearly outweighs previous SOTA finetuning-free methods on all of the metrics, including subject alignment, image-text alignment, subject exclusivity, and overall quality. Remarkably, it also outperforms fine-tuning-based methods in image quality and image-text alignment within both benchmarks. Particularly in the Multi-Subject Benchmark, the SSR-Encoder demonstrates outstanding performance in subject exclusivity, markedly outperforming competing methods. This highlights the efficacy of its selective representation capability and editability. While Dreambooth excels in subject alignment within the DreamBench dataset, the SSR-Encoder and Break-A-Scene show comparable performance on the Multi-Subject Bench. This suggests that although Dreambooth is highly effective in capturing detailed subject information, SSR-Encoder achieves a balanced and competitive performance in subject representation.

\textbf{Qualitative comparison}.
Fig.~\ref{image1} displays the high-fidelity outcomes produced by the SSR-Encoder using diverse query inputs, affirming its robustness and zero-shot generative capabilities. The SSR-Encoder demonstrates proficiency in recognizing and focusing on common concepts, ensuring an accurate representation of the selected image subjects. Its seamless integration with other customized models and control modules further solidifies its significant role in the stable diffusion ecosystem.

In qualitative comparisons, as depicted in Fig.~\ref{compare}, Textual Inversion and Reference Only encounter difficulties in maintaining subject identity. Dreambooth, IP-adapter, and BLIP-Diffusion, although advanced, exhibit limitations in effectively disentangling intertwined subjects. Break-A-Scene achieves commendable subject preservation but at the cost of extensive fine-tuning. ELITE, with its focus on local aspects through masks, also faces challenges in consistent identity preservation.

In contrast, our SSR-Encoder method stands out for its fast generation of selected subjects while adeptly preserving their identities. This capability highlights the method's superior performance in generating precise and high-quality subject-driven images, thereby addressing key challenges faced by other current methods.

\begin{figure}
\includegraphics[width=0.48\textwidth]{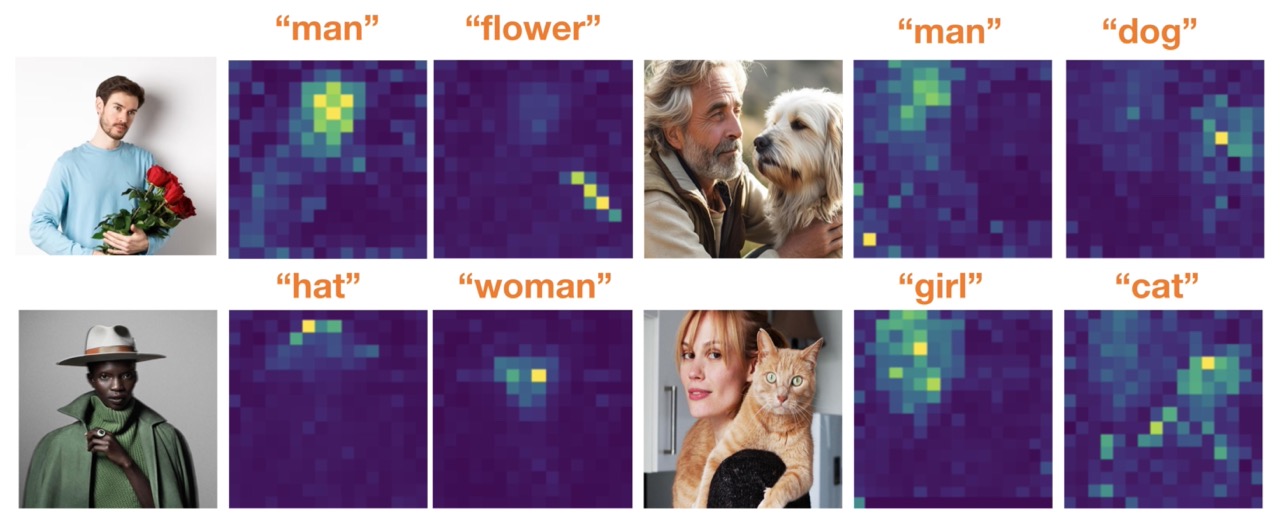}
    \caption{Visualization of attention maps $A_{t2p}$.}
    \vspace{-0.4cm}
    \label{image8}
\end{figure}

\textbf{Ablation study.}
Our ablation study begins with visualizing the attention maps generated by our Token-to-Patch Aligner, as shown in Fig.~\ref{image8}. These maps demonstrate how different text tokens align with corresponding patches in the reference image, evidencing the Aligner's effectiveness.

\begin{figure}
\centering
\includegraphics[width=0.48\textwidth]{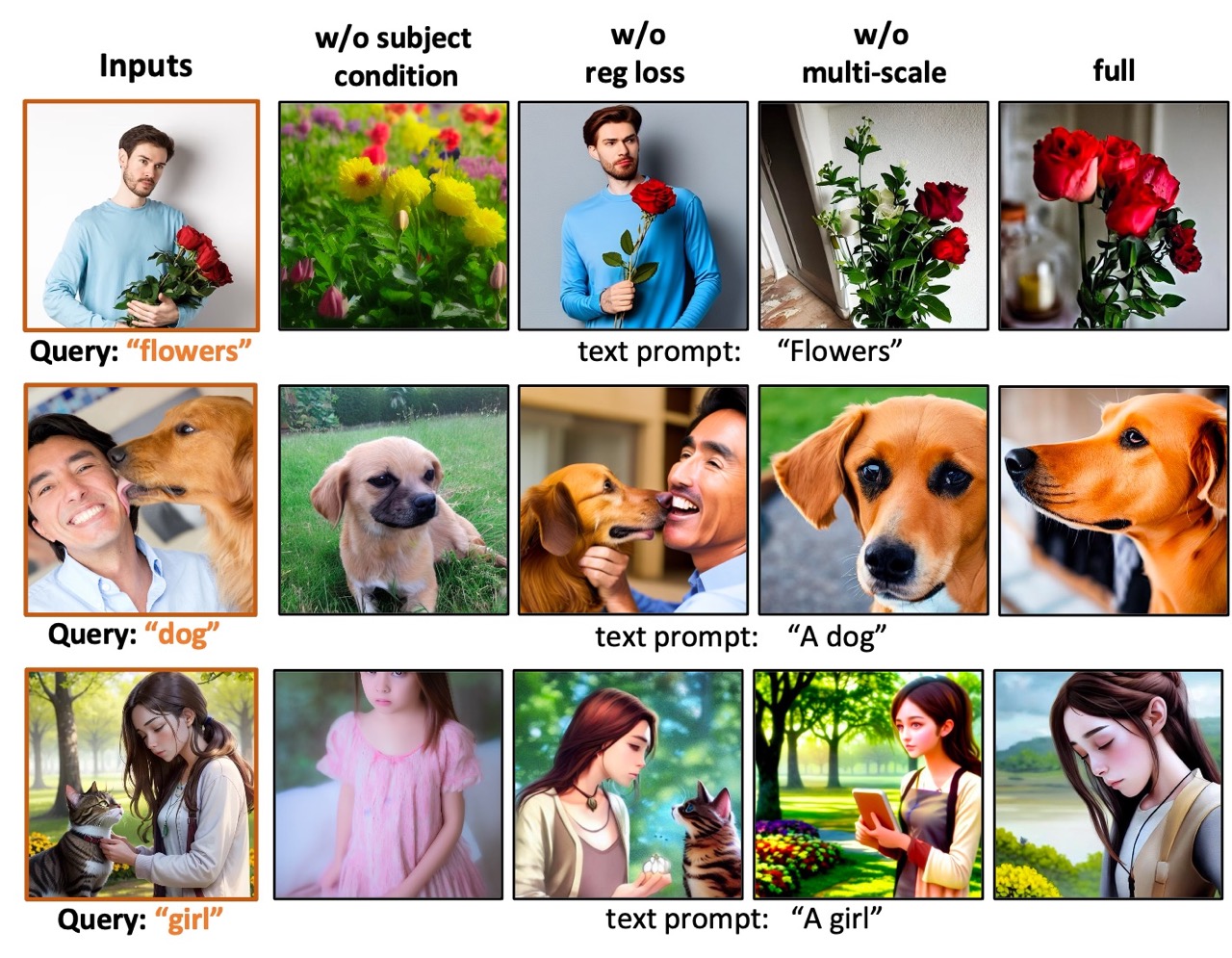}
    \caption{\textbf{Qualitative ablation.} We ablate our approach by using different model settings. Without the $L_{reg}$, the model struggles to exclude undesired subjects from reference images. Substituting the multi-scale image feature results in less detailed outputs.}
    \vspace{-0.4cm}
    \label{image4}
\end{figure}

To evaluate the significance of various components, we conducted experiments by systematically removing them and observing the outcomes. Initially, we removed the subject condition, relying solely on the text condition for image generation, to determine if the subject details could be implicitly recalled by the base model. Subsequently, we trained a model without the embedding consistency regularization loss ($L_{reg}$) to assess its criticality. We also substituted our multi-scale visual embedding with a conventional last-layer visual embedding. The results of these experiments are depicted in Fig.~\ref{image4}.

\setlength{\tabcolsep}{2mm}{
\begin{table}
\begin{footnotesize}
\centering
\caption{\textbf{Ablation results on Multi-subject Bench.} Removing each component would lead to a performance drop on different aspects.}
\label{tab3}
\begin{tabular}{c|ccc}
\toprule
{\small Ablation Setups} & \small CLIP-T $\uparrow$ & \small CLIP-ES $\downarrow$  & \small DINO-M $\uparrow$ \\
\midrule
Text2Image &\textbf{0.352} & -- & 0.318  \\
Ours(w/o multi-scale) &0.257 & 0.185 & 0.510\\
Ours(w/o reg loss) &0.235 & 0.199 & 0.552 \\
Ours(full) &0.302 & \textbf{0.182} &\textbf{0.556}\\
\bottomrule
\end{tabular}

\end{footnotesize}
\vspace{-0.4cm}
\end{table}
}

Our observations reveal that without subject conditioning, the generated subjects failed to correspond with the reference image. Omitting the multi-scale image feature resulted in a loss of detailed information, as evidenced by a significant drop in the DINO-M score. Discarding the embedding consistency regularization loss led to challenges in generating specific subjects from coexisting subjects, adversely affecting the CLIP-ES score. In contrast, the full implementation of our method demonstrated enhanced expressiveness and precision.

Quantitative comparisons, as shown in Table~\ref{tab3}, also indicate that our complete method achieves the best results across subject exclusivity and subject alignment. It slightly trails the original Stable Diffusion (SD) model only in text-image alignment.  Substituting the multi-scale visual embedding significantly impacts image consistency, while excluding the embedding consistency regularization loss hampers text-image consistency.

 \section{Conclusion}
In this paper, we introduced the SSR-Encoder, a groundbreaking finetuning-free approach for selective subject-driven image generation. This method marks a significant advancement in the field, offering capabilities previously unattainable in selective subject representation. At its core, the SSR-Encoder consists of two pivotal the token-to-patch aligner and the detail-preserving subject encoder. The token-to-patch aligner effectively aligns query input tokens with corresponding patches in the reference image, while the subject encoder is adept at extracting multi-scale subject embeddings, capturing fine details across different scales. Additionally, the incorporation of a newly proposed embedding consistency regularization loss further enhances the overall performance of the system. Our extensive experiments validate the SSR-Encoder's robustness and versatility across a diverse array of scenarios. The results clearly demonstrate the encoder's efficacy in generating high-quality, subject-specific images, underscoring its potential as a valuable tool in the open-source ecosystem.

\clearpage
{
    \small

    \bibliographystyle{IEEEtran}
}

\setcounter{page}{1}
\maketitlesupplementary
\renewcommand\thesection{\Alph{section}}
In this supplementary material, we first introduce the preliminaries of Diffusion and CLIP in Section~\ref{preliminary}. Following that, we provide an in-depth discussion on our Detail-Preserving Image Encoder in Section~\ref{encoder}. In subsequent sections, we introduce the methods we compared against and the user study we conducted, specifically in Section~\ref{sce_compare} and Section~\ref{userstudy} respectively. We also present our results on human image generation in Section~\ref{sce_face}. Additional results from our work on Dreambench and Multi-subject bench are showcased in Section~\ref{results}. We then provide further details about our training data and Multi-subject bench in Section~\ref{data}. In Section~\ref{sce_cn} and Section~\ref{sce_animate}, we present the outcomes generated by combining our SSR-encoder with ControlNet~\cite{controlnet} and animatediff~\cite{animatediff}, which not only demonstrates the generalization of our SSR encoder but also illustrates its seamless applicability in the realm of controllable generation and video generation for maintaining character consistency with reference images. Lastly, we analyze the broader impact brought by our method and the limitation of our method in Section~\ref{impact} and Section~\ref{lim}.

\section{Preliminaries}
\label{preliminary}
\subsection{Preliminary for Diffusion Models}
Diffusion Model (DM)~\cite{ddpm, scoremodel} belongs to the category of generative models that denoise from a Gaussian prior $\mathbf{x_T}$ to target data distribution $\mathbf{x_0}$ by means of an iterative denoising procedure. The common loss used in DM is:
\begin{equation}
    L_{simple}(\bm{\theta}) := \mathbb{E}_{\mathbf{x_0}, t, \bm{\epsilon}}\left[\left\|\bm{\epsilon}-\bm{\epsilon_\theta}\left(\mathbf{x_t}, t\right)\right\|_2^2\right],
\end{equation}
where $\mathbf{x_t}$ is an noisy image constructed by adding noise $\bm{\epsilon} \in \mathcal{N}(\mathbf{0},\mathbf{1})$  to the natural image $\mathbf{x_0}$ and the network $\bm{\epsilon_\theta(\cdot)}$ is trained to predict the added noise. At inference time, data samples can be generated from Gaussian noise $\bm{\epsilon} \in \mathcal{N}(\mathbf{0},\mathbf{1})$ using the predicted noise $\bm{\epsilon_\theta}(\mathbf{x_t}, t)$ at each timestep $t$ with samplers like DDPM~\cite{ddpm} or DDIM~\cite{ddim}.

Latent Diffusion Model (LDM)~\cite{ldm} is proposed to model image representations in autoencoder’s latent space. LDM significantly speeds up the sampling process and facilitates text-to-image generation by incorporating additional text conditions. The LDM loss is:
\begin{equation}
    L_{LDM}(\bm{\theta}) := \mathbb{E}_{\mathbf{x_0}, t, \bm{\epsilon}}\left[\left\|\bm{\epsilon}-\bm{\epsilon_\theta}\left(\mathbf{x_t}, t, \bm{\tau_{\theta}}(\mathbf{c})\right)\right\|_2^2\right],
\end{equation}
where $\mathbf{x_0}$ represents image latents and $\bm{\tau_\theta(\cdot)}$ refers to the BERT text encoder~\cite{bert} used to encodes text description $\mathbf{c_t}$.

Stable Diffusion (SD) is a widely adopted text-to-image diffusion model based on LDM. Compared to LDM, SD is trained on a large LAION~\cite{laion5b} dataset and replaces BERT with the pre-trained CLIP~\cite{clip} text encoder.

\subsection{Preliminary for CLIP}
CLIP~\cite{clip} consists of two integral components: an image encoder represented as $F(x)$, and a text encoder, represented as $G(t)$. The image encoder, $F(x)$, transforms an image $x$ with dimensions $\mathbb{R}^{3 \times H \times W}$ (height $H$ and width $W$) into a $d$-dimensional image feature $f_x$ with dimensions $\mathbb{R}^{N \times d}$, where $N$ is the number of divided patches. On the other hand, the text encoder, $G(t)$, creates a $d$-dimensional text representation gt with dimensions $\mathbb{R}^{M \times d}$ from natural language text $t$, where $M$ is the number of text prompts. Both encoders are concurrently trained using a contrastive loss function that enhances the cosine similarity of matched pairs while reducing that of unmatched pairs. After training, CLIP can be applied directly for zero-shot image recognition without the need for fine-tuning the entire model.

\section{Designing Choice of Image Encoder}
\label{encoder}
In this section, we conduct a preliminary reconstruction experiment to demonstrate that vanilla image features fail to capture fine-grained representations of the target subject and verify the effectiveness of our method. We first introduce our experimental setup and evaluation metrics in Sec.~\ref{b-setup}. Subsequently, we explain the implementation details of each setting in Sec.~\ref{b-implementation}. Finally, we conduct qualitative and quantitative experiments in Sec.~\ref{b-result} to prove the superiority of our proposed methods compared to previous works.

\subsection{Experimental Setup}
\label{b-setup}

In our image reconstruction experiment, we investigate four types of image features. The details are as shown in Fig.~\ref{fig:secB_framework}:

\begin{figure}[!h]
    \centering
    \includegraphics[width=1\linewidth]{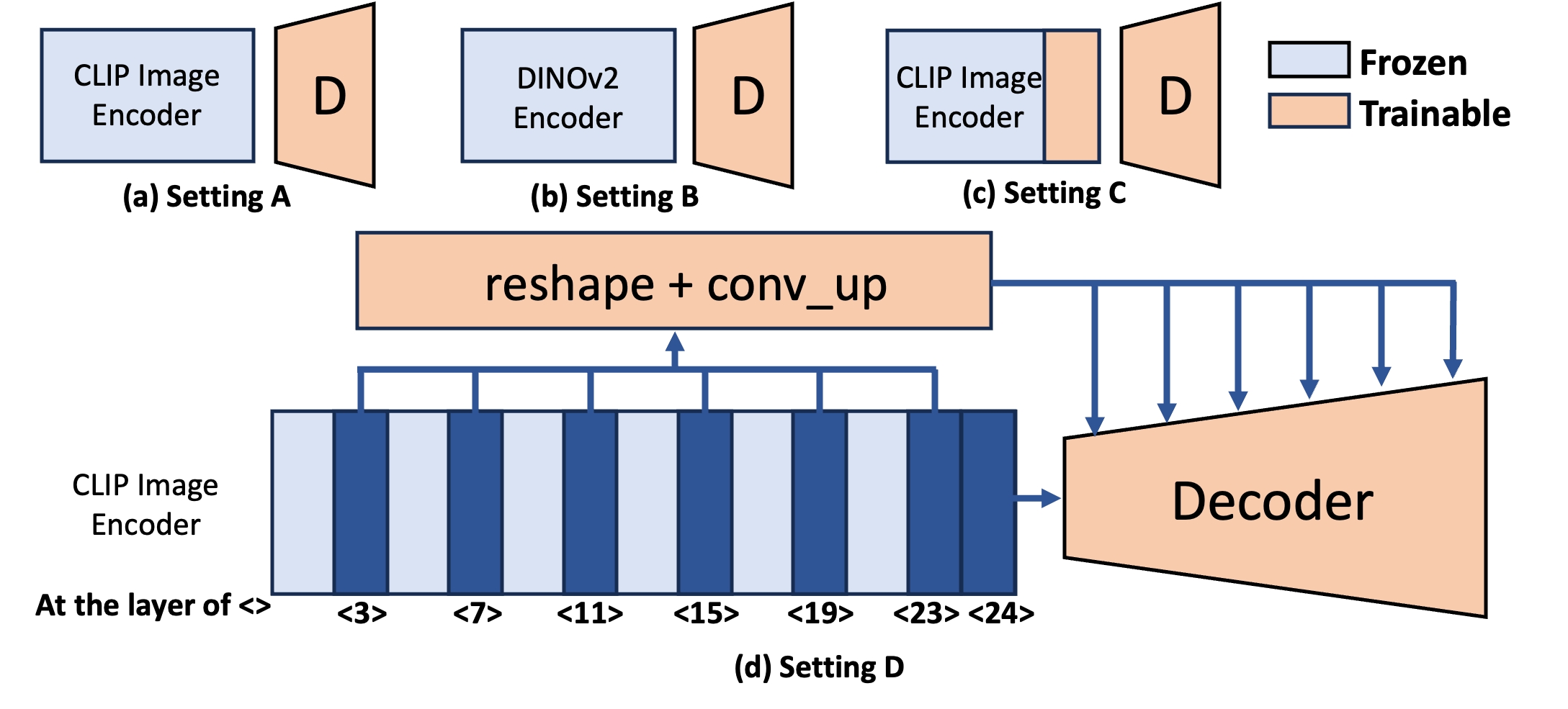}
    \caption{Details for each setting.}
    \label{fig:secB_framework}
    \vspace{-0.4cm}
\end{figure}

\begin{itemize}
    \item \textbf{Setting A: CLIP Image Features.} In this setting, we employ the vanilla CLIP image encoder to encode the input image and utilize the features from the final layer as the primary representation for subsequent reconstruction.
    \item \textbf{Setting B: DINOv2 Image Features.} Analogous to setting A, we replace the CLIP image encoder with the DINOv2 encoder to extract the features.    
    \item \textbf{Setting C: Fine-tuned CLIP Image Features.} With the goal of recovering more fine-grained details while preserving text-image alignment, we fine-tune the last layer parameters of the CLIP image encoder using a CLIP regularization loss.
    \item \textbf{Setting D: Multi-scale CLIP Image Features.} Instead of fine-tuning, we resort to using features from different scales of the CLIP backbone as the image representations.
\end{itemize}

To verify the effectiveness of our methods, we employ the following metrics: \textbf{Perceptual Similarity (PS)}~\cite{perceptual_similarity} and \textbf{Peak Signal-to-Noise Ratio (PSNR)} to assess the quality of reconstruction, \textbf{CLIP-T}~\cite{clipscore} and \textbf{Zero-Shot ImageNet Accuracy (ZS)}~\cite{imagenet} to access the preservation of text-image alignment in image encoder variants.

As for data used in our preliminary experiments, we utilize a subset of LAION-5B~\cite{laion5b}. This dataset comprises approximately 150,000 text-image pairs for training and a further 10,000 text-image pairs designated for testing. 

\subsection{Implementation Details}
\label{b-implementation}
We use OpenCLIP ViT-L/14~\cite{openclip} and DINOv2 ViT-L/14~\cite{dinov2} as the image encoders and all images are resized to 224$\times$224 for training. 
The model underwent 100,000 training iterations on 4 V100 GPUs, using a batch size of 32 per GPU.  
We adopt the Adam optimizer~\cite{adam} with a learning rate of 3e-4 and implement the one-cycle learning scheduler. 
To better preserve the pre-trained weights, we set the learning rate of the image encoder as 1/10 of the other parameters if fine-tuning is required.
We adopt the same architecture of the VAE decoder in LDM~\cite{ldm} with an extra upsampling block and employ nearest interpolation to obtain the final reconstruction results.
We adopt $L_2$ reconstruction loss in all our settings and additionally employ $L_{clip}$ when fine-tuning the CLIP encoder.

\subsection{Experiment Results} 
\label{b-result}

\begin{figure}[!h]
    \centering
    \includegraphics[width=1.0\linewidth]{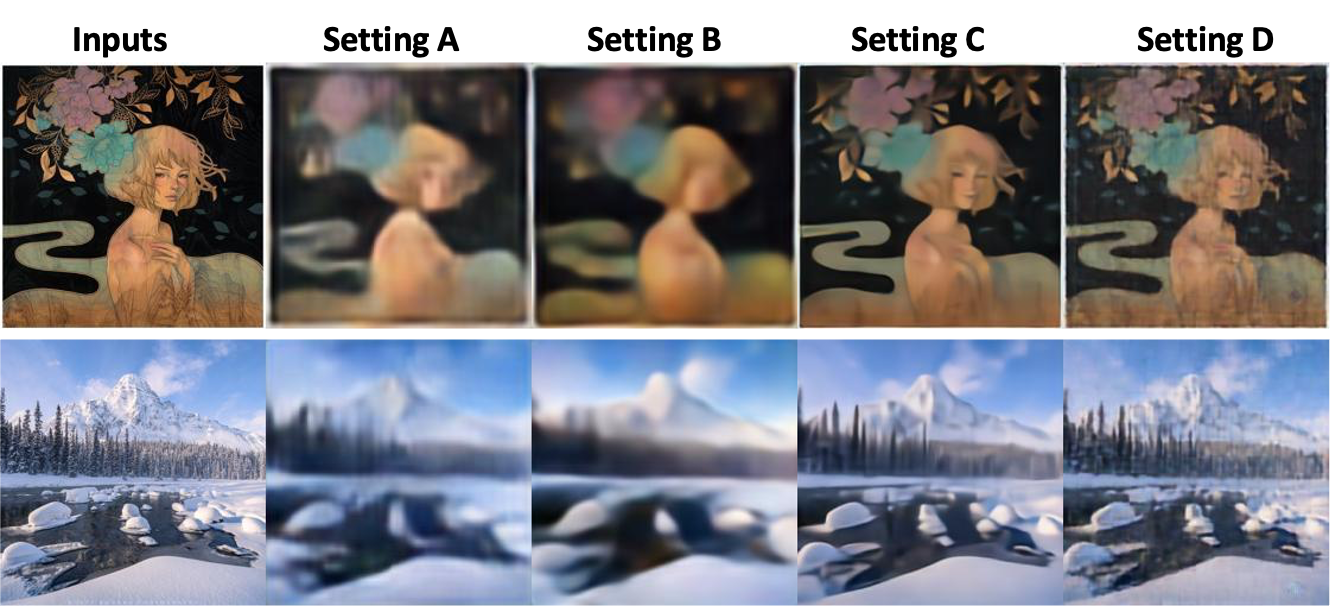}
    \caption{Comparisons of different settings.}
    \label{fig:secB_qualitative}
    \vspace{-0.4cm}
\end{figure}

\textbf{Qualitative results.} 
To demonstrate the effectiveness of our method, we present reconstruction results in Fig.~\ref{fig:secB_qualitative}. It is observed that vanilla CLIP image features and DINOv2 features only result in rather blurry outcomes. By contrast, both fine-tuned CLIP image features and multi-scale CLIP image features manage to retain more details. Specifically, multi-scale CLIP image features is able to generate sharp edges without obvious degradations. Consequently, we infer that multi-scale features are more competent at preserving the fine-grained details we require.

\textbf{Quantitative results.} 
The quantitative results are shown in Table~\ref{tab_subB_quantitative}. In terms of reconstruction quality, it's noteworthy that both the fine-tuned CLIP image features and multi-scale CLIP image features are adept at producing superior outcomes, exhibiting lower perceptual similarity scores and higher PSNR. This indicates that these features are more representative than either vanilla CLIP image features or DINOv2 features. However, despite the assistance from CLIP regularization loss, fine-tuned CLIP image features still suffer significant degradation in text-image alignment, which fails to meet our requirements. Consequently, we opt for multi-scale features as our primary method for extracting subject representation.

\setlength{\tabcolsep}{3mm}{
\begin{table}[!h]
\begin{footnotesize}
\centering
\caption{Comparisons of different settings.}
\label{tab_subB_quantitative}
\begin{tabular}{c|cccc}
\toprule
{\small Settings} & \small PS $\downarrow$ & \small PSNR $\uparrow$ & \small CLIP-T $\uparrow$  & \small ZS $\uparrow$ \\
\midrule
A &     0.0036           & 28.63             & \textbf{0.1816}   & \textbf{75.3\%}  \\
B &     0.0013           & 28.56             & --               & -- \\
C &     \textbf{0.0004}  & \textbf{29.73}    & 0.1394           & 68.4\% \\
D &     0.0006           & 29.49             & \textbf{0.1816}  & \textbf{75.3\%} \\
\bottomrule
\end{tabular}
\end{footnotesize}
\vspace{-0.4cm}
\end{table}
}

\begin{figure*}[!h]
    \centering
\includegraphics[width=1\linewidth]{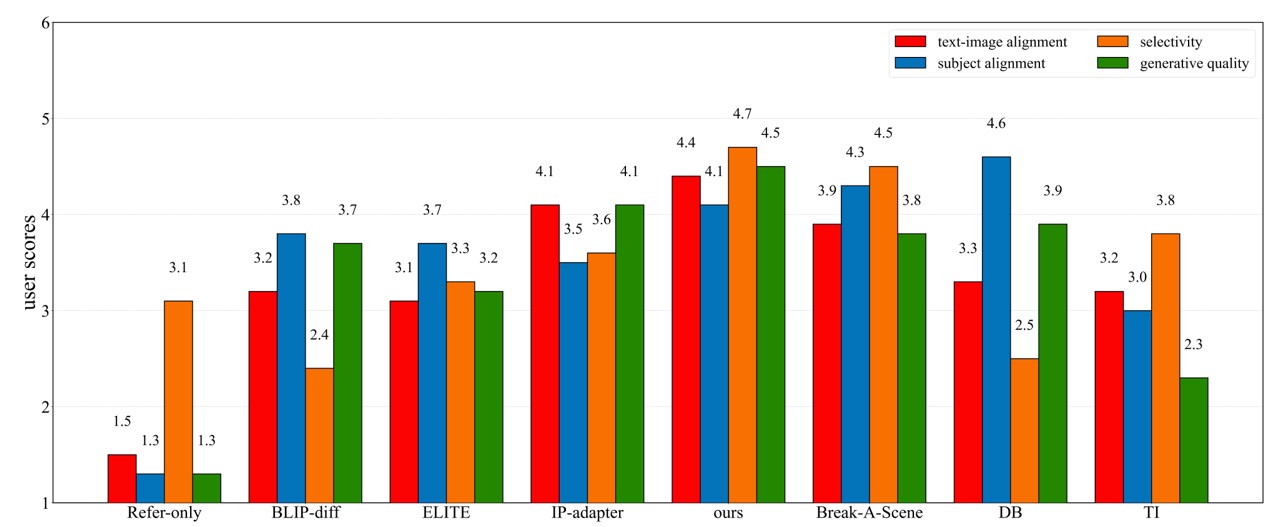}
    \caption{User study comparisons of different methods.}
    \label{fig:enter-label}
\end{figure*}

\section{Details of Comparison Experiments}
\label{sce_compare}

\subsection{Details of Compared Methods}

\begin{enumerate}
    \item \textbf{Finetune-based Methods:}
    \begin{itemize}
        \item \textbf{Textual Inversion} \cite{TI}: A method to generate specific subjects by describing them using new ``words" in the embedding space of pre-trained text-to-image models.
        \item \textbf{Dreambooth} \cite{DB}:  A method of personalized image generation by fine-tuning the parameters in diffusion U-Net structure.
        \item \textbf{Break-A-Scene} \cite{breakascene}: Aims to extract a distinct text token for each subject in a single image, enabling fine-grained control over the generated scenes.
    \end{itemize}

    \item \textbf{Finetune-free Methods:}
    \begin{itemize}
        \item \textbf{Reference only} \cite{Mikubill2022}: Guide the diffusion directly using images as references without training through simple feature injection.
        \item \textbf{ELITE} \cite{elite}: An encoder-based approach encodes the visual concept into the textual embeddings for subject-driven image generation.
        \item \textbf{IP-adapter} \cite{ye2023ip}: Focuses on injecting image information without fine-tuning the base model.
        \item \textbf{BLIPDiffusion} \cite{blipdiffusion}: Combines BLIP's language-image pretraining with diffusion models.
    \end{itemize}
\end{enumerate}

These methods were chosen for their relevance and advancements in the field, providing a robust frame of reference for evaluating the performance and innovations of our SSR-Encoder. 

\subsection{Details of Implementation}

In order to achieve a fair comparison, all the methods are implemented using the official open-source code based on SD v1-5 and the official recommended parameters. For the Multi-subject bench, all the methods use a single image as input and utilize different subjects to guide the generation. We provide 6 different text prompts for each subject on each image and generate 6 images for each text prompt. For Dreambench, we follow~\cite{DB,blipdiffusion} and generate 6 images for each text prompt provided by DreamBench.

\section{User Study}
\label{userstudy}
We conducted a user study to compare our method with DB, TI, Break-A-Scene, ELITE, and IP-adapter perceptually. For each evaluation, each user will see one input image with multiple concepts, two different prompts for different concepts, and 5 images generated by each prompt and each method. 60 evaluators were asked to rank each generated image from 1 (worst) to 5 (best) concerning its selectivity, text-image alignment, subject alignment, and generative quality. The results are shown in Table.~\ref{fig:enter-label} indicate that our method outperforms the comparison methods in generative quality and better balances subject consistency and text-image alignment.

\section{Human Image Generation}
\label{sce_face}
\begin{figure*}[!h]
    \centering
    \includegraphics[width=1.0\linewidth]{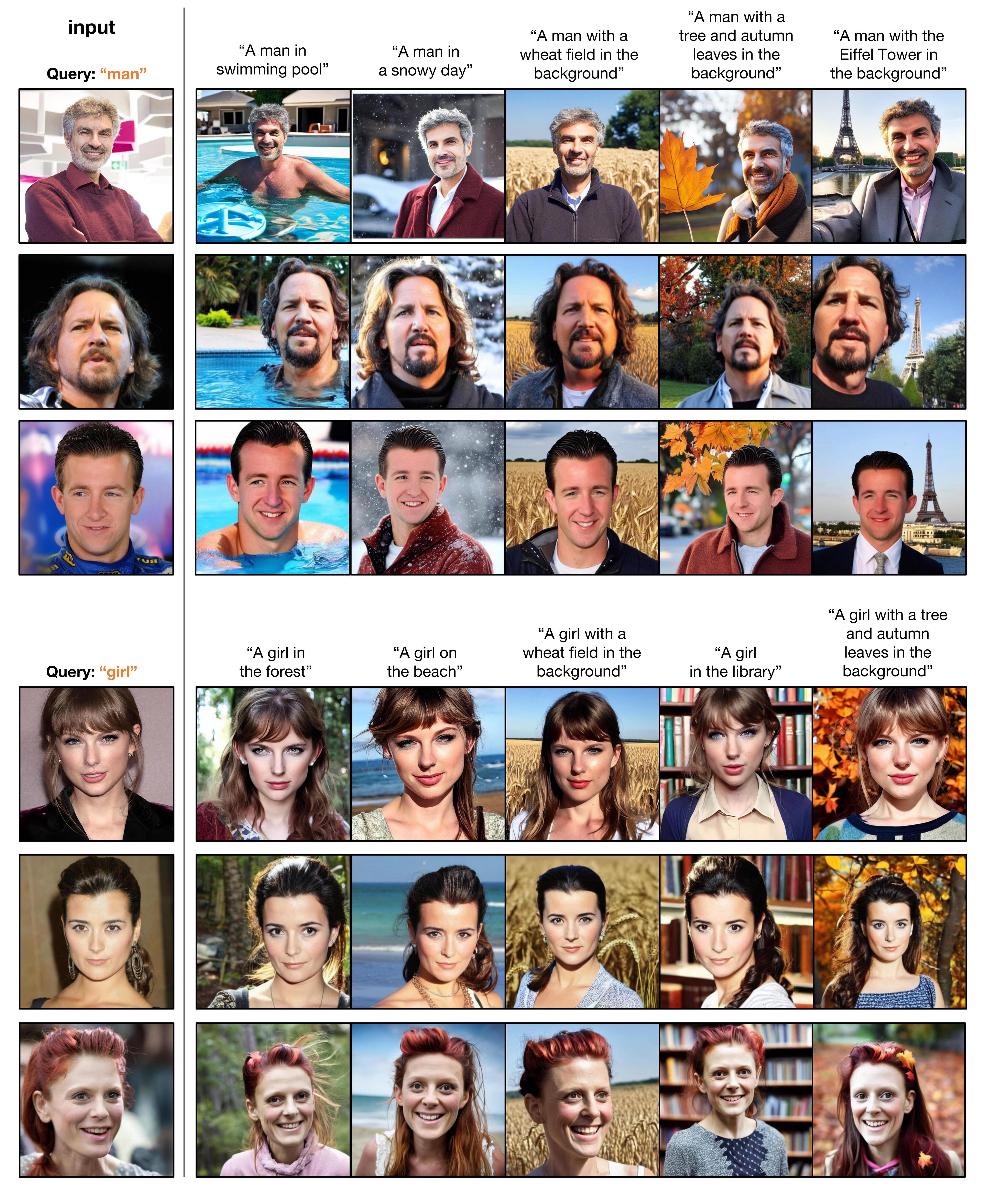}
    \caption{Results for human image generation.}
    \label{face}
    \vspace{-0.4cm}
\end{figure*}

Despite the SSR-Encoder not being trained in domain-specific settings (such as human faces), it is already capable of capturing the intricate details of the subjects. For instance, similar to the method outlined in~\cite{subject}, we utilize face images from the OpenImages dataset~\cite{open} as reference images for generating human images. Fig.~\ref{face} showcases samples of the face images we generated. To better illustrate our results, we also employ images of two celebrities as references.

\section{Ablations of $\tau$ and $\lambda$}
\label{sce_ab}
\begin{figure}[ht]
    \centering
    \includegraphics[width=1.0\linewidth]{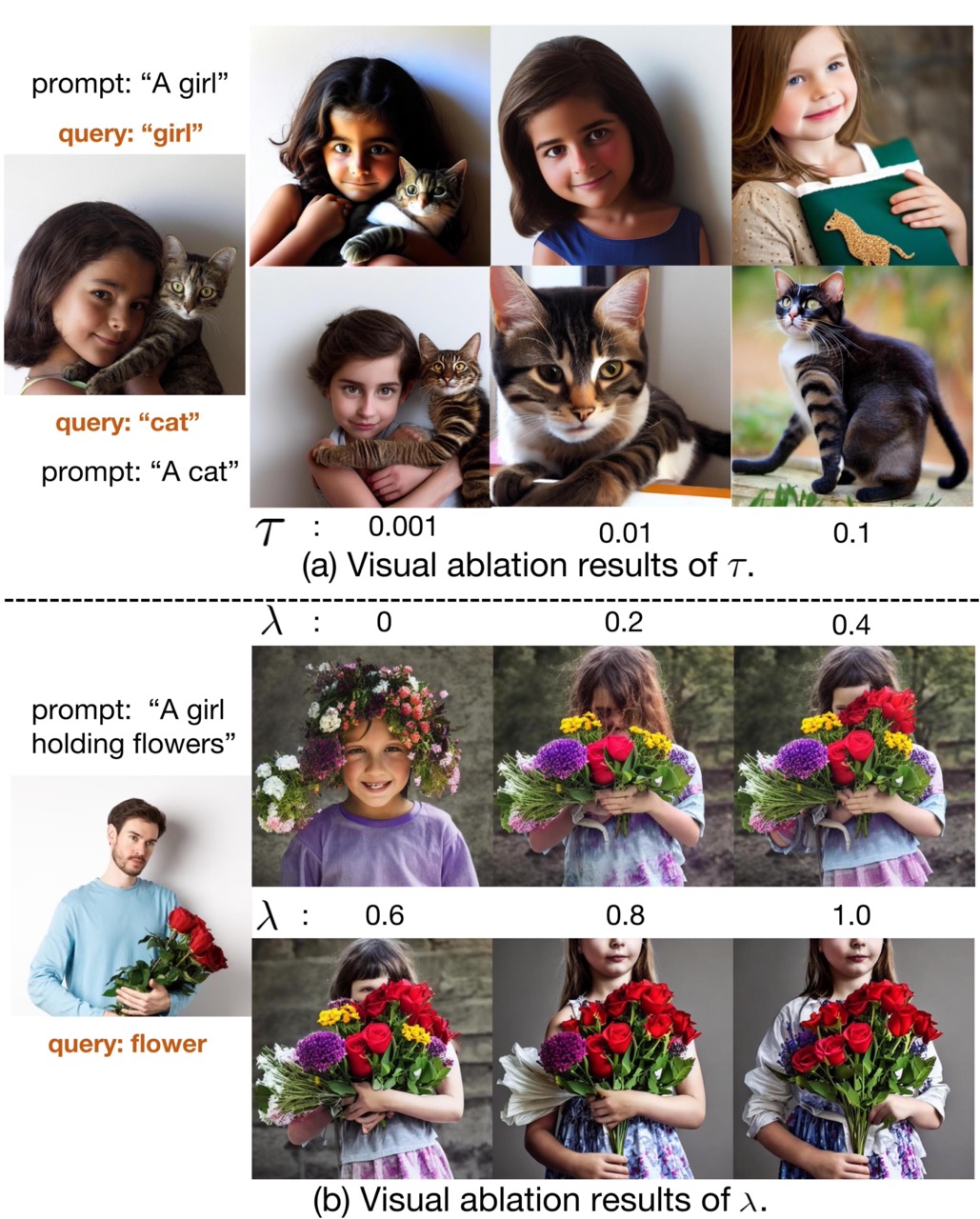}
    \caption{Visual ablation results of $\tau$ and $\lambda$.}
    \label{sup_ab}
    \vspace{-0.4cm}
\end{figure}

As shown in Fig.~\ref{sup_ab} (a), under the same training settings, when $\tau$ was 0.01, the model managed to balance both identity consistency and selectivity. The effects of different $\lambda$ values on the images under ablation and fixed seed conditions are shown in Fig.~\ref{sup_ab} (b). The smaller $\lambda$, the weaker the influence of the reference image.

\section{Examples of Evaluation Samples}
\label{results}
\begin{figure*}[!hp]
\vspace{-1cm}
    \centering
    \includegraphics[width=0.8\linewidth]{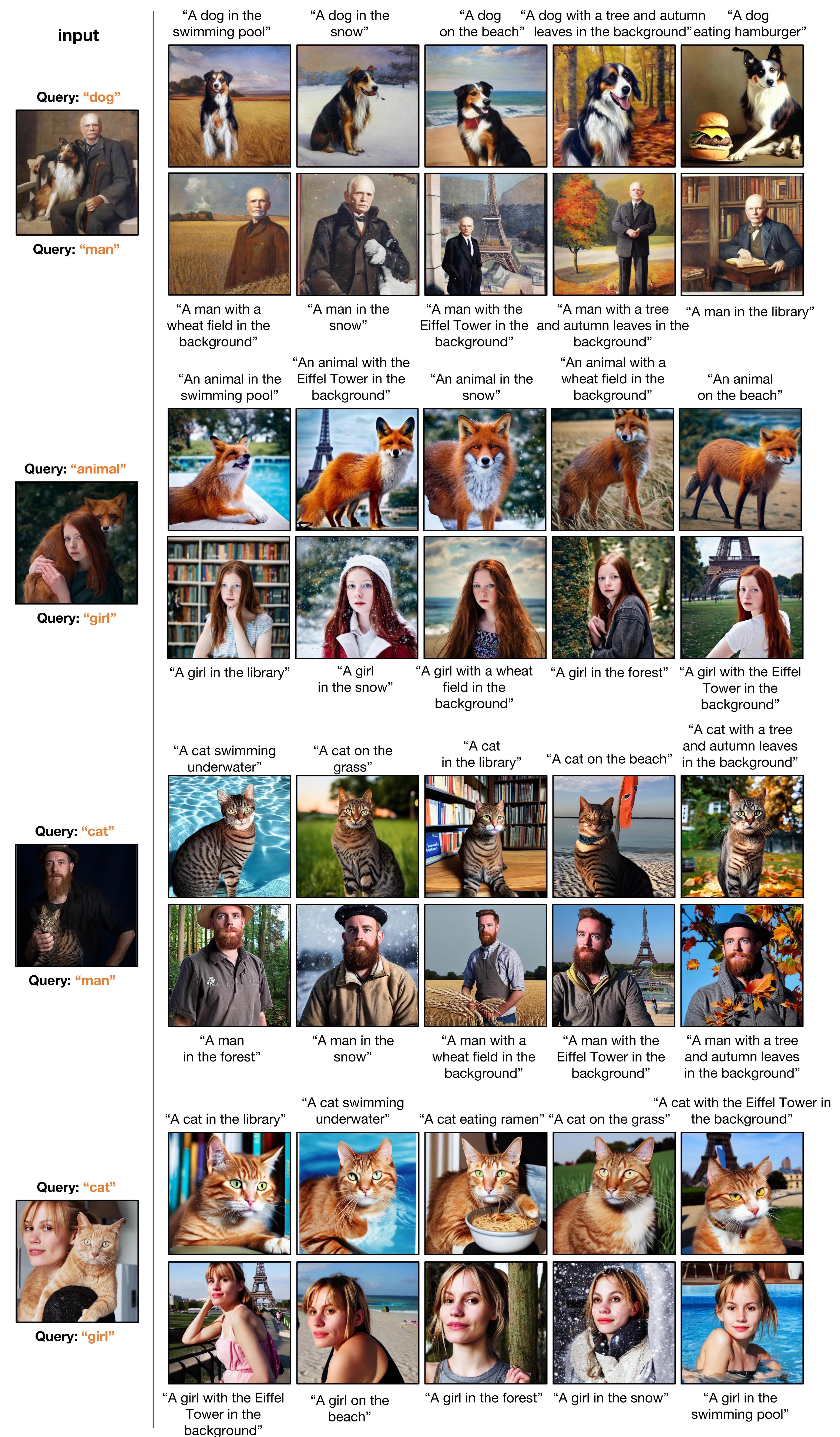}
    \caption{Examples of evaluation samples on the multi-subject bench.}
    \label{multi1}

\end{figure*}

\begin{figure*}[!hp]
\vspace{-1cm}
    \centering
    \includegraphics[width=0.8\linewidth]{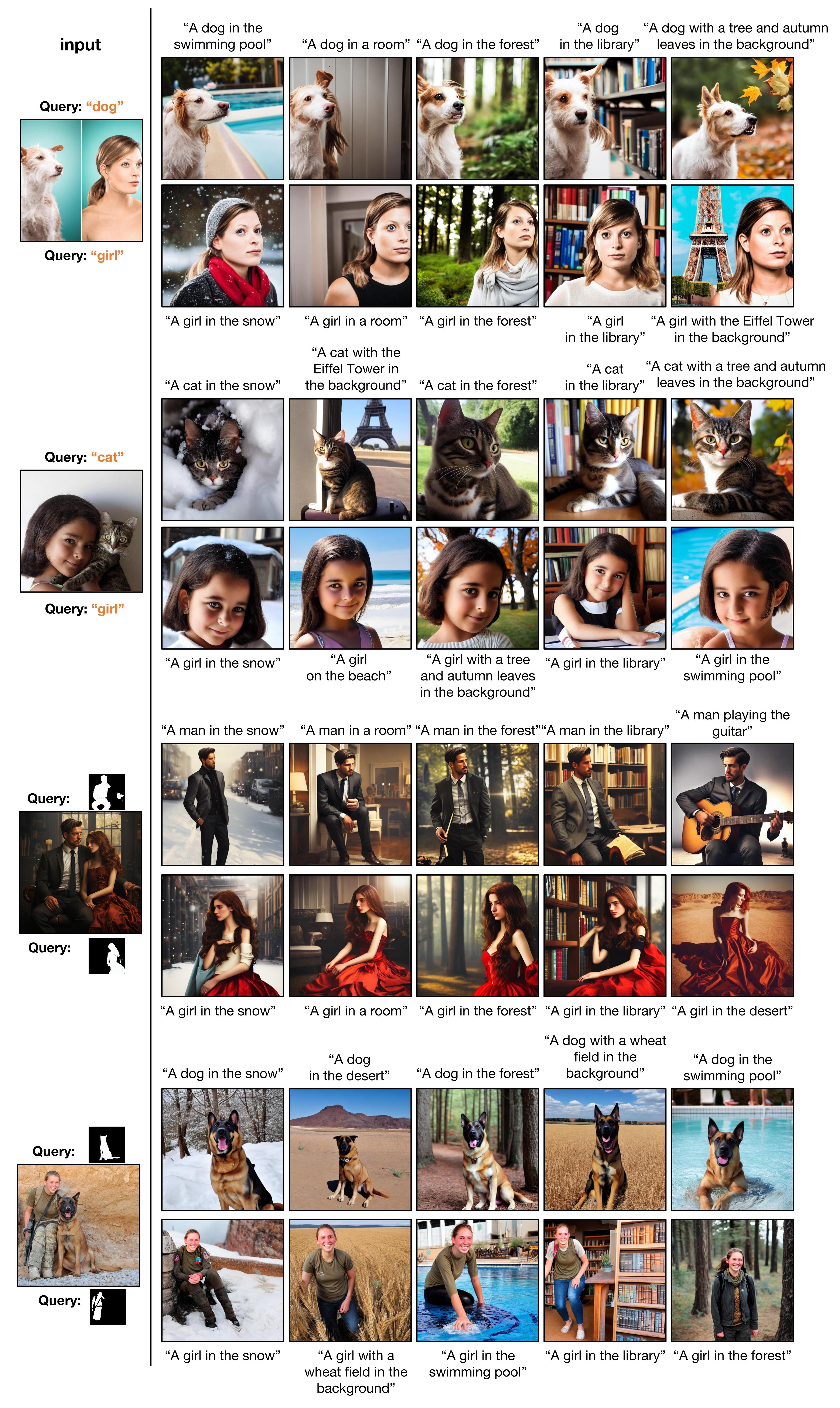}
    \caption{Examples of evaluation samples on the multi-subject bench.}
    \label{multi2}

\end{figure*}

\begin{figure*}[!hp]
\vspace{-0.7cm}
    \centering
    \includegraphics[width=0.8\linewidth]{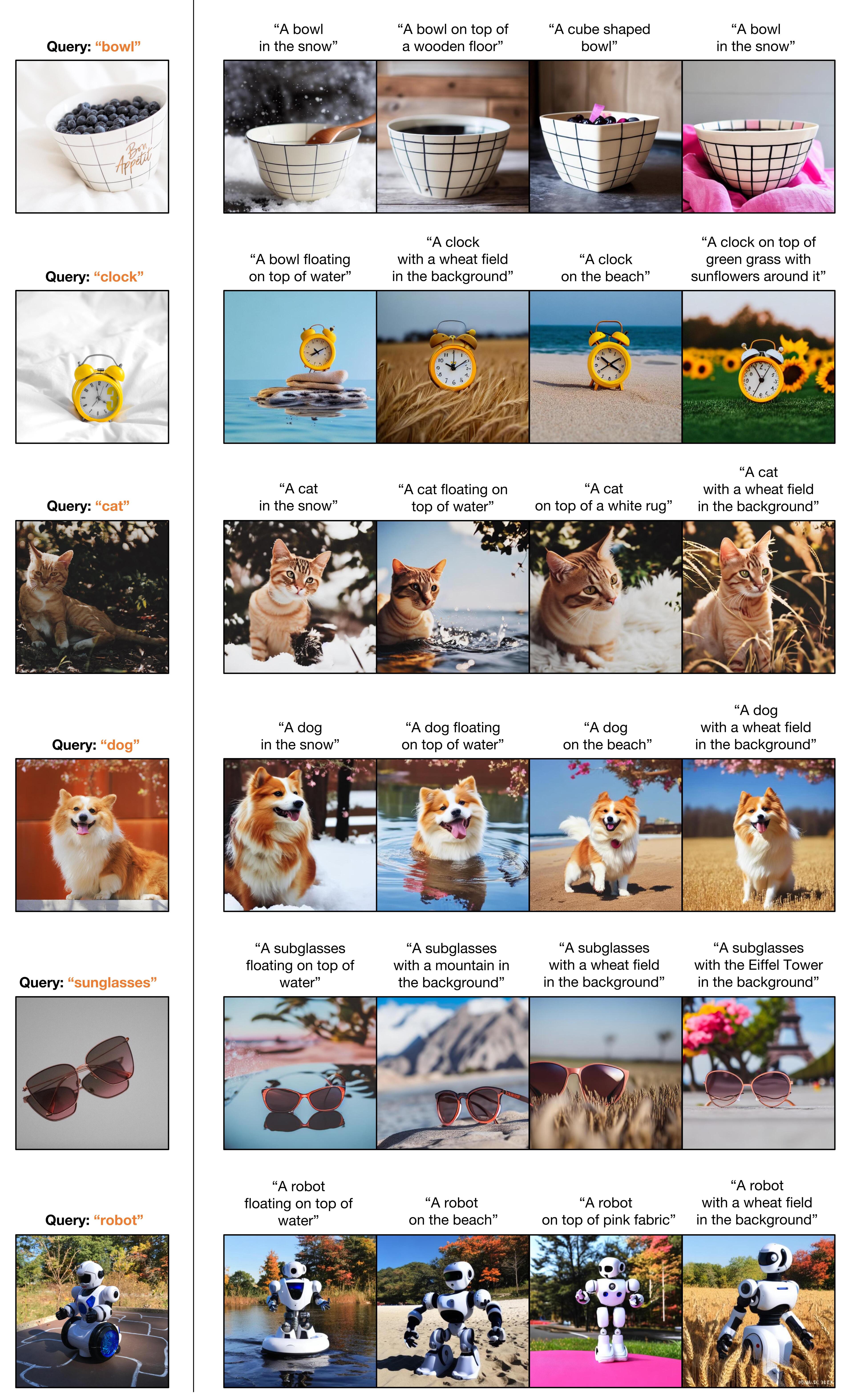}
    \caption{Examples of evaluation samples on the dreambench.}
    \label{dream}

\end{figure*}

In this section, we present more evaluation samples in our method on two different test datasets: Multi-Subject bench and DreamBench bench in Fig.~\ref{multi1}, Fig.~\ref{multi2}, and Fig.~\ref{dream}. 

Moreover, we present more qualitative comparison results in Fig.~\ref{compare}. As illustrated in the figure, our approach is more adept at focusing on the representation of distinct subjects within a single image, utilizing a query to select the necessary representation. In contrast to other methods, our method does not result in ambiguous subject extraction, a common issue in finetune-based methods. For instance, in the Dreambooth row from Fig.~\ref{compare}, two subjects frequently appear concurrently, indicating a low level of selectivity. When considering selectivity, generative quality, and text-image alignment, our SSR-Encoder surpasses all methods and achieves the level of finetune-based methods in terms of subject alignment.

\section{Details of Our Training Data and the Multi-subject Bench}
\label{data}
\begin{itemize}
    \item \textbf{Details of training data.} 
    Our model utilizes the Laion 5B dataset\cite{laion5b}, selecting images with aesthetic scores above 6.0. The text prompts are re-captioned using BLIP2~\cite{blip2}. The dataset comprises 10 million high-quality image-text pairs, with 5,000 images reserved for testing and the remainder for training.
    Clearly, the distribution of training data has a significant impact on our model. The more a particular type of subject data appears in the training data capt, the better our performance on that type of subject. Therefore, we further analyze the word frequency in the training data caption and report the most frequent subject descriptors in the table\ref{tab_word}.
    
\setlength{\tabcolsep}{1.6mm}{
\begin{table}[!h]
\begin{footnotesize}
\centering
\caption{The most frequent subject descriptors in our training data.}
\label{tab_word}
\begin{tabular}{cc|cc|cc}
\toprule
{\small Subject} & \small frequency & \small Subject & \small frequency  & \small subject &frequency \\

\midrule

 woman&1528518 & suit&256732 &dog &164819 \\
 man&1256613 & trees&240771 & snow&163838 \\
 people&536434 & hair&229538 & girl &162311 \\
 table&385643 &wooden &216958 &hat &157549 \\
 mountain&315765 &street &212259 &flowers &152308 \\
 chairs&291189 & house&191785 & sky&151332 \\
 dress&268058 &building &168670 &cat &147851 \\

\bottomrule
\end{tabular}
\end{footnotesize}
\vspace{-0.4cm}
\end{table}
}

    \item \textbf{Details of multi-subject bench.} 
The Multi-subject Bench comprises 100 images from our test data. More specifically, the data is curated based on the caption associated with each image from our test set. An image progresses to the next stage if its caption contains at least two subject descriptors. Subsequently, we verify the congruence between the caption and the image. If the image aligns with the caption and adheres to human aesthetic standards, it is shortlisted as a candidate image. Ultimately, we meticulously selected 100 images from these candidates to constitute the Multi-subject Bench.

\end{itemize}

\section{Compatibility with ControlNet}
\label{sce_cn}
\begin{figure*}[!hp]
\vspace{-0.7cm}
    \centering
    \includegraphics[width=1.0\linewidth]{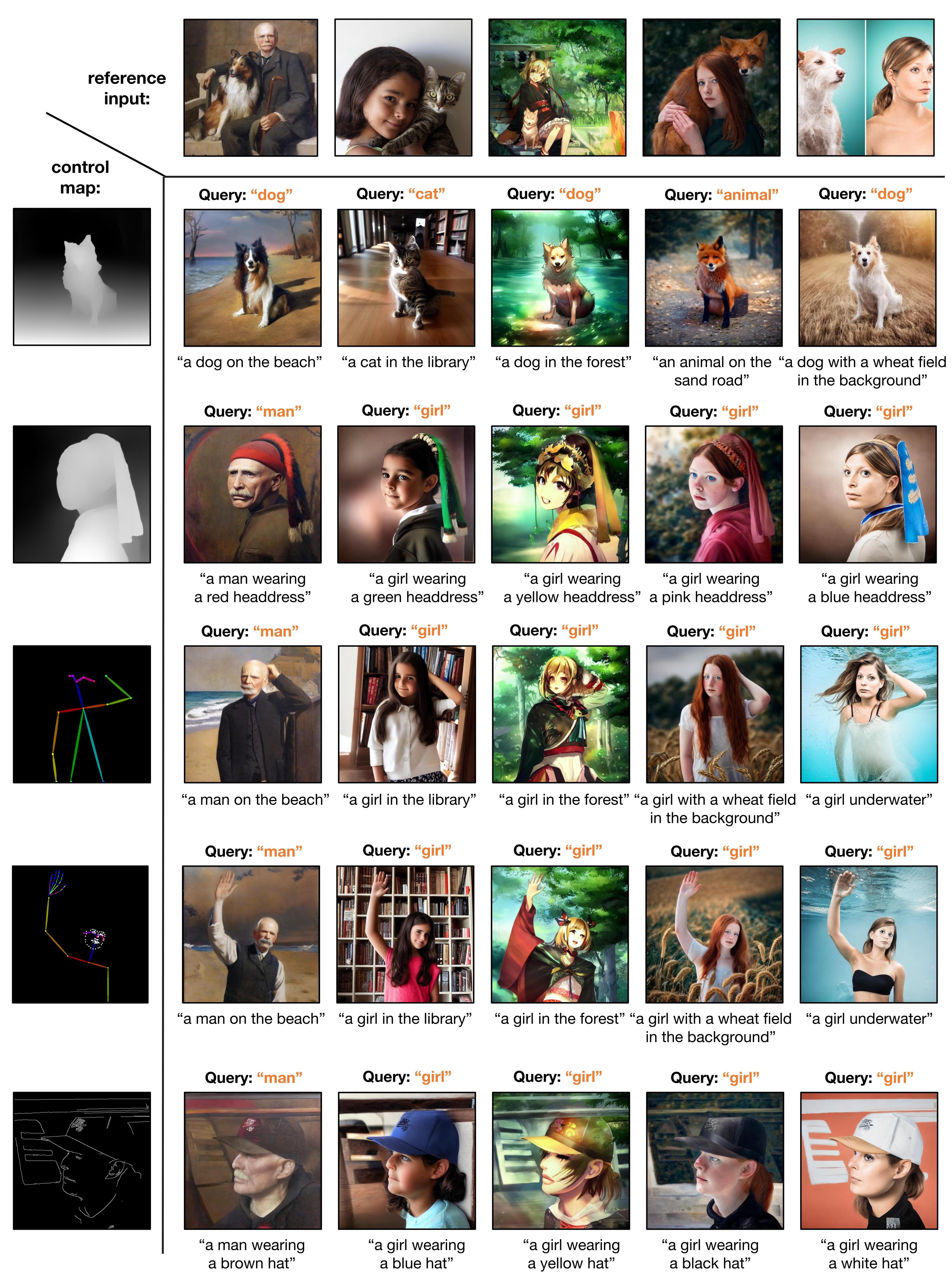}
    \caption{Results of combining our SSR-Encoder with controlnet.}
    \label{cn}

\end{figure*}

Our SSR-Encoder can be efficiently integrated into controllability modules. As demonstrated in Fig.~\ref{cn}, we present additional results of amalgamating our SSR-Encoder with ControlNet~\cite{controlnet}. Our approach can seamlessly merge with controllability modules, thereby generating controllable images that preserve consistent character identities in alignment with reference images.

\section{Compatibility with AnimateDiff}
\label{sce_animate}
\begin{figure*}[!hp]
\vspace{-0.7cm}
    \centering
    \includegraphics[width=1.05\linewidth]{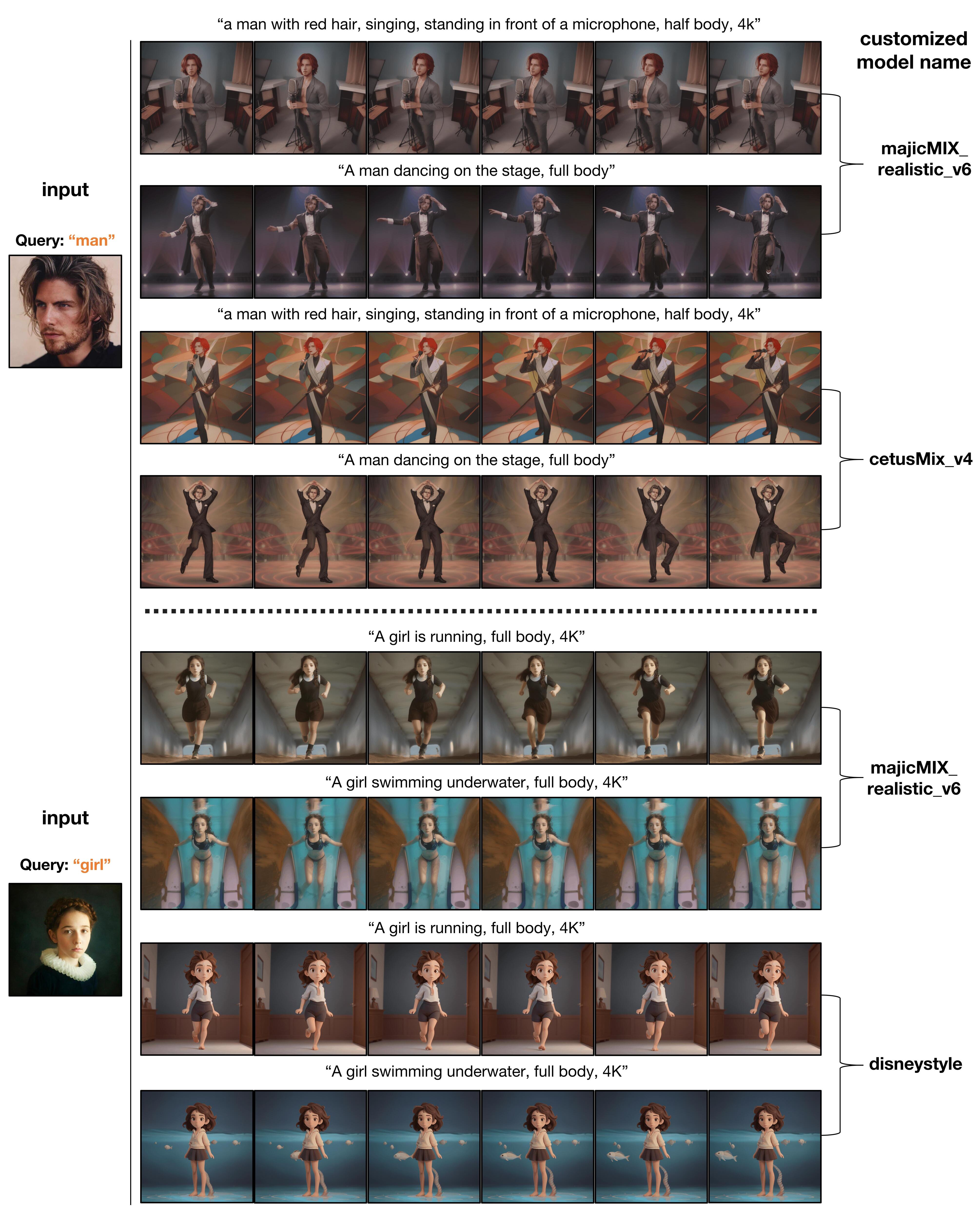}
    \caption{Results of combining our SSR-Encoder with Animatediff.}
    \label{animate}

\end{figure*}

Our SSR-Encoder is not only versatile enough to adapt to various custom models and controllability modules, but it can also be effectively applied to video generation, integrating seamlessly with video generation models. In Fig.~\ref{animate}, we demonstrate the impact of combining our SSR-Encoder with Animatediff~\cite{animatediff}. Despite not being trained on video data, our method can flawlessly combine with Animatediff to produce videos that maintain consistent character identities with reference images.

\section{Broader Impact}
\label{impact}
Our method in subject-driven image generation holds significant potential for advancing the field of text-to-image generation, particularly in creating personalized images. This technology can be applied across various domains such as personalized advertising, artistic creation, and game design, and can enhance research at the intersection of computer vision and natural language processing. However, while the technology has numerous positive applications, it also raises ethical and legal considerations. For instance, generating personalized images using others' images without appropriate permission could infringe upon their privacy and intellectual property rights. Therefore, adherence to relevant ethical and legal guidelines is crucial. Furthermore, our model may generate biased or inappropriate content if misused. We strongly advise against using our model in user-facing applications without a thorough inspection of its output and recommend proper content moderation and regulation to prevent undesirable consequences.

\section{Limitation}
\label{lim}
Due to the uneven distribution of the filtered training data, we found that the fidelity will be slightly worse for some concepts that are uncommon in our training data. This can be addressed by increasing the training data. We plan to address these limitations and extend our approach to 3D generation in our future work.

\begin{figure*}[!hp]
\vspace{-0.7cm}
    \centering
    \includegraphics[width=0.9\linewidth]{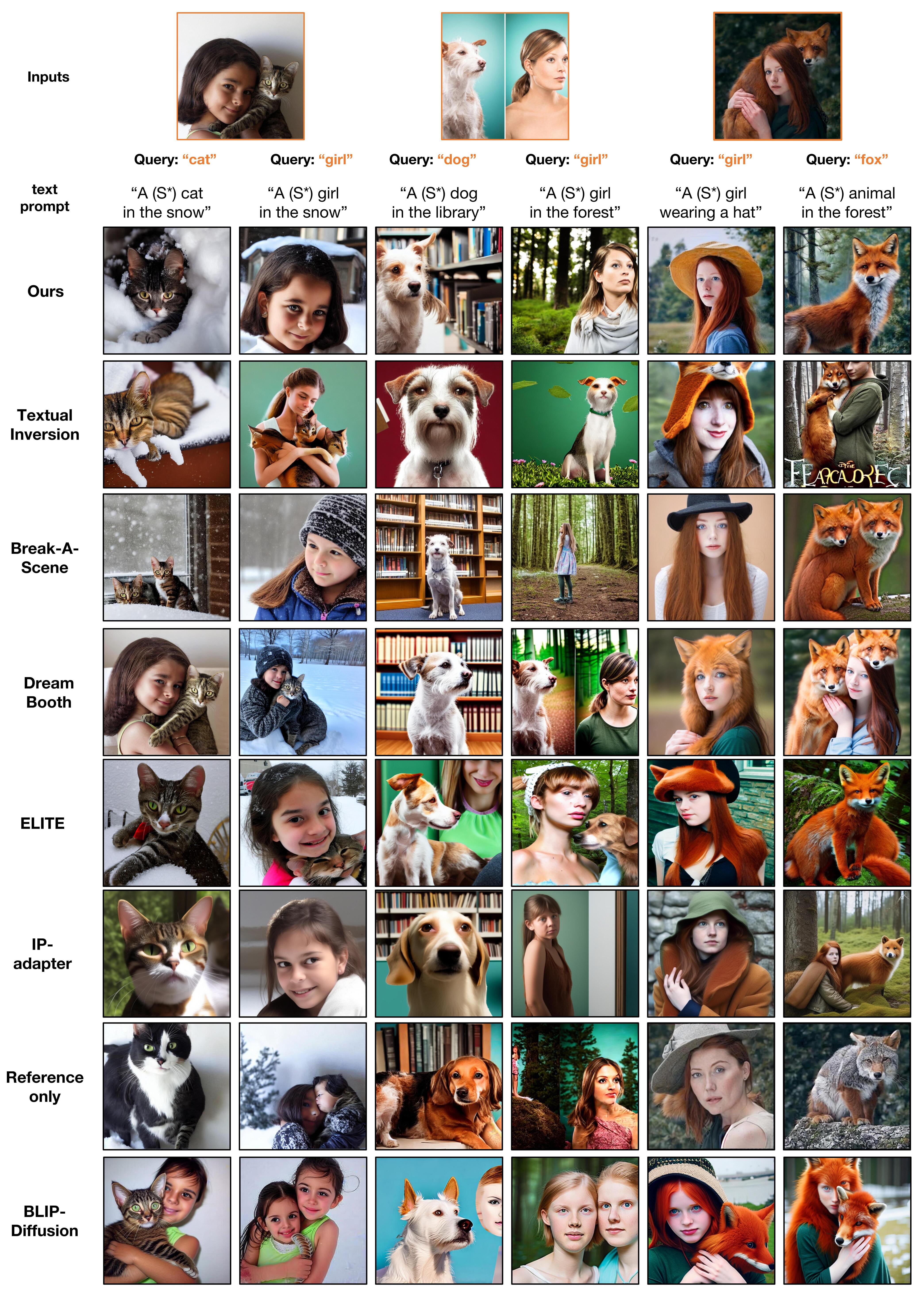}
    \caption{More results of the qualitative comparison.}
    \label{compare}
\end{figure*}

\end{document}